\documentclass[nohyperref]{article}

\usepackage{microtype}
\usepackage{graphicx}
\usepackage{subfigure}
\usepackage{booktabs} % for professional tables
\usepackage{url}
\usepackage{amsmath,bm}
\usepackage{mathrsfs}
\usepackage{multirow}
\usepackage{CJKutf8}
\usepackage{verbatim}
\usepackage{amssymb}
\usepackage{xspace}
\usepackage{enumitem}
\usepackage{makecell}
\usepackage{ragged2e}

\usepackage{hyperref}

\usepackage[accepted]{icml2022}

\usepackage{amsmath}
\usepackage{amssymb}
\usepackage{mathtools}
\usepackage{amsthm}
\usepackage[bottom]{footmisc}

\usepackage[capitalize,noabbrev]{cleveref}

\theoremstyle{plain}

\theoremstyle{definition}

\theoremstyle{remark}

\usepackage[textsize=tiny]{todonotes}

\newcommand{\E}{{\mathbb{E}}}

\newcommand{\LL}{{\mathcal{L}}}
\newcommand{\Lmain}{\LL}
\newcommand{\LpathA}{\LL_{A}}
\newcommand{\Lmp}{\LL_{\text{glancing}}}
\newcommand{\bv}{\mathbf{v}}
\newcommand{\bg}{\mathbf{g}}
\newcommand{\R}{\mathbb{R}}
\DeclareMathOperator*{\argmax}{arg\,max}

\newcommand{\methodname}{DA-Transformer\xspace}
\newcommand{\graphname}{DAG\xspace}
\newcommand{\graphnames}{DAGs\xspace}

\newcommand{\tempneg}{}
\newcommand{\stdfont}[1]{\scalebox{.6}{#1}}

\definecolor{mypurple}{HTML}{7030A0}
\definecolor{mygreen}{HTML}{385723}
\definecolor{myorange}{HTML}{ED7D31}
\definecolor{myblue}{HTML}{2F5597}
\definecolor{mygray}{HTML}{7F7F7F}

\newcommand{\revise}[1]{#1}
\newcommand{\rrevise}[1]{#1}
\newcommand{\rrrevise}[1]{#1}
\newcommand{\rrrrevise}[1]{#1}

\icmltitlerunning{Directed Acyclic Transformer for Non-Autoregressive Machine Translation}

\DeclareUrlCommand\Code{\urlstyle{rm}}
\expandafter\def\expandafter\UrlBreaks\expandafter{\UrlBreaks  
\do\/\do\a\do\b\do\c\do\d\do\e\do\f\do\g\do\h\do\i\do\j\do\k
\do\l\do\m\do\n\do\o\do\p\do\q\do\r\do\s\do\t\do\u\do\v
\do\w\do\x\do\y\do\z
\do\A\do\B\do\C\do\D\do\E\do\F\do\G\do\H\do\I\do\J\do\K
\do\L\do\M\do\N\do\O\do\P\do\Q\do\R\do\S\do\T\do\U\do\V
\do\W\do\X\do\Y\do\Z}

\begin{document}
\begin{CJK}{UTF8}{gbsn}

\twocolumn[
\icmltitle{Directed Acyclic Transformer for Non-Autoregressive Machine Translation}



\icmlsetsymbol{equal}{*}

\begin{icmlauthorlist}
\icmlauthor{Fei Huang}{coai,tsinghua,equal}
\icmlauthor{Hao Zhou}{bytedance}
\icmlauthor{Yang Liu}{tsinghua}
\icmlauthor{Hang Li}{bytedance}
\icmlauthor{Minlie Huang}{coai,tsinghua} \\
\end{icmlauthorlist}

\icmlaffiliation{coai}{The CoAI group, Tsinghua University, China.}
\icmlaffiliation{tsinghua}{Institute for Artificial Intelligence, State Key Lab of Intelligent Technology and Systems, Beijing National Research Center for Information Science and Technology, Department of Computer Science and Technology, Tsinghua University, China.}
\icmlaffiliation{bytedance}{ByteDance AI Lab}

\icmlcorrespondingauthor{Fei Huang}{\Code{f-huang18@mails.tsinghua.edu.cn}}
\icmlcorrespondingauthor{Hao Zhou}{\Code{haozhou0806@gmail.com}}
\icmlcorrespondingauthor{Yang Liu}{\Code{liuyang2011@tsinghua.edu.cn}}
\icmlcorrespondingauthor{Hang Li}{\Code{lihang.lh@bytedance.com}}
\icmlcorrespondingauthor{Minlie Huang}{\Code{aihuang@tsinghua.edu.cn}}

\icmlkeywords{Non-autoregressive Text Generation, Machine Translation}

\vskip 0.3in
]



\printAffiliationsAndNotice{}{\textsuperscript{*}This work is partially done during Fei Huang's internship at ByteDance AI Lab.}  % leave blank if no need to mention equal contribution

\begin{abstract}
Non-autoregressive Transformers (NATs) significantly reduce the decoding latency by generating all tokens in parallel. However, such independent predictions prevent NATs from capturing the dependencies between the tokens for generating multiple possible translations.
\rrevise{In this paper, we propose Directed Acyclic Transfomer (\methodname), which represents the hidden states in a Directed Acyclic Graph (\graphname), where each path of the DAG corresponds to a specific translation. 
The whole \graphname simultaneously captures multiple translations and facilitates fast predictions in a non-autoregressive fashion.}
\rrrevise{Experiments on the raw training data of WMT benchmark show that \methodname substantially outperforms previous NATs by about 3 BLEU on average, which is the first NAT model that achieves competitive results with autoregressive Transformers without relying on knowledge distillation.}
\end{abstract}

\section{Introduction}
\label{sec:intro}
\addtolength{\skip\footins}{-0.2em}
\addtolength{\abovedisplayskip}{-0.15em}
\addtolength{\belowdisplayskip}{-0.15em}

\begin{figure}[!t]
    \centering
    \includegraphics[width=1\linewidth]{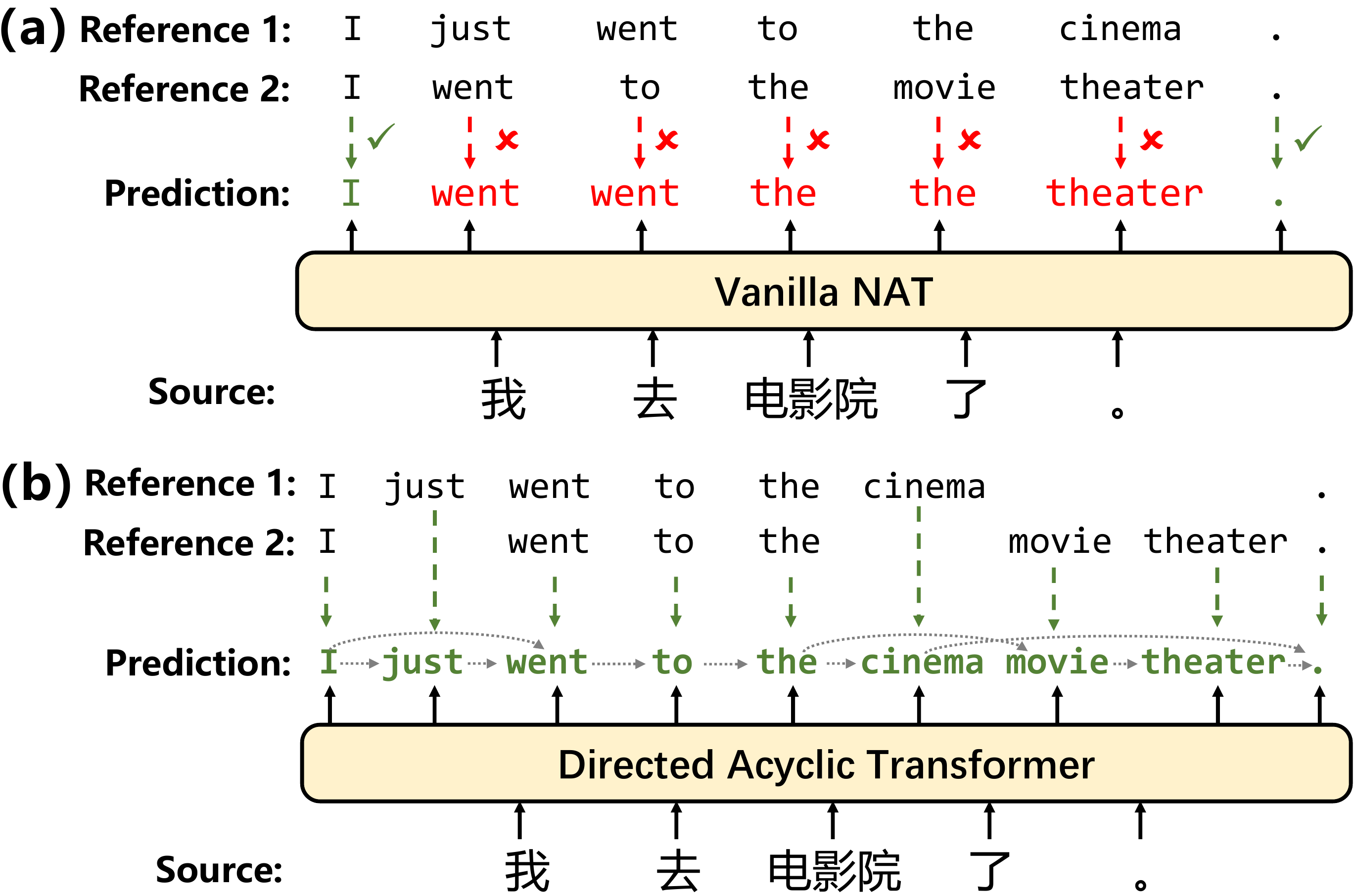}
    \vspace{-1.8em}
    \caption{(a) The multi-modality problem in vanilla NAT. Multiple possible references provide inconsistent labels at some positions, leading to an implausible prediction that mixes \rrrevise{several} translations. (b) The proposed \textit{\methodname}. \rrevise{Tokens from different \rrrevise{translations} are assigned to distinct vertices to avoid the inconsistent labels.} In inference, we follow the transitions to recover \rrrevise{the output translation}.}
    \label{fig:intro}
    \vspace{-1.1em}
\end{figure}

Transformer has been the most popular architecture for seq\-uence-to-sequence learning, especially for machine translation~\cite{transformer2017vaswani}.
Vanilla Transformer adopts the autoregressive approach for generation, which obtains strong results but is inefficient in inference due to its sequential decoding.
\revise{To tackle the problem, Non-autoregressive Transformers (NATs, \citealp{nat2018gu}; \citealp{levenshtein2019gu}; \citealp{flowseq2019ma};\citealp{lexical2021ding}; \citealp{tricktrade2021gu}) have been proposed}, which significantly reduce the inference latency by predicting all tokens in parallel and achieve reasonably high performances in translation. \rrrevise{Notably, an NAT-based system obtain the highest BLEU score in German to English translation of WMT21~\cite{qian-etal-2021-volctrans,akhbardeh-etal-2021-findings}, even better than a line of Autoregressive Transformer~(AT) systems. }

However, current NATs severely suffer from the \textit{multi-modality problem}~\cite{nat2018gu} in both training and inference.\footnote{The \textit{multi-modality} here refers to the fact that there are multiple possible translations for a single source sentence.}
Intuitively, in training, as shown in Fig.\ref{fig:intro}(a), NAT models are trained to predict each token independently, where one position may have several possible tokens as labels \rrrevise{from several different translation references.}
In such a case, an NAT model may learn to generate an implausible output mixing multiple translations.
Additionally, in inference, the NAT still cannot sample fluent translations even if it captures multi-modal information in training.
Since the NAT model generates all tokens simultaneously, no effective sampling approach can be used on top of it.
In contrast, ATs do not have the same problem because of their left-to-right generation, where the multi-modality problem for a later position is not so severe \rrrevise{since} its prefix has been given.

\rrrevise{Currently, the main solution to} address the multi-modality problem is to reduce the data modalities by knowledge distillation~(KD, \citealp{seqkd2016kim}; \citealp{nat2018gu}), namely, replacing the original training targets with predicted sentences from an AT teacher.
KD is simple yet effective, \rrrevise{which always leads to significant BLEU improvements, e.g., about 8 BLEU points on WMT14 En-De for vanilla NATs.}

However, we argue that current state-of-art NAT models heavily rely on KD, which has two crucial disadvantages.
a) Training NAT models by distilling from AT makes the training process redundant. We need to train an AT model first and then regenerate the whole training data. Such complex pre-processing prevents NATs from being practically used. 
b) Generally, the student model in KD cannot outperform its teacher model with a large margin. In such a case, KD restricts NAT's performance by imposing an upper bound~(\textit{not strict}), which seriously hurts the potential of further developing NAT models.

In this paper, we propose \textit{Directed Acyclic Transformer}~(\methodname) for Non-Autoregressive Machine Translation, which directly captures many translation modalities via a proposed \textit{Directed Acyclic Decoder}, instead of indirectly reducing modalities by KD.
\revise{Specifically, different from decoders of ATs or vanilla NATs, our proposed decoder organizes the hidden states as a Directed Acyclic Graph~(DAG) rather than a sequence.
As shown in Fig.\ref{fig:intro}(b), the DAG has multiple paths, each of which corresponds to a specific sentence.}\footnote{The DAG is similar to the concept of word lattice~\cite{DBLP:conf/icassp/RichardsonOR95}. The words are represented by edges instead of vertices in the word lattice, and in contrast, each vertex of the DAG in our model represents a word distribution rather than concrete words.} 
In training, \rrevise{the DAG structure} enables \methodname to capture \rrevise{multiple} translation modalities simultaneously, which avoids the inconsistent labels in vanilla NAT training.
In inference, it can generate sentences along predicted paths, which not only avoids incorrect outputs mixing multiple translations but also enables the generation of diverse translations by sampling different paths.

Notice that \methodname predicts all translation words in parallel, and the whole model is trained in an end-to-end fashion, which enjoys all merits of NAT models.
We propose an objective that does not require multiple references in training, making it applicable to most translation benchmarks.
In inference, we propose several sampling methods to decode a translation from \methodname, which provides flexible quality-latency tradeoff in generation. 

Experimental results show that \methodname significantly reduces the gap between NATs and ATs while preserving the inference latency (7x $\sim$ 14x speedup over ATs).
Especially on WMT17 Zh-En, our best model outperforms autoregressive Transformer by 0.6 BLEU without the help of knowledge distillation.
To our best knowledge, it is the first time that a \textit{non-iterative} NAT model achieves competitive results with AT models without KD.
\methodname outperforms existing NATs~(including iterative approaches) with a large margin on the raw data of standard En$\leftrightarrow$DE and En$\leftrightarrow$Zh benchmarks, which sufficiently shows the effectiveness of our proposed model.

\section{Related Work}

\textbf{Non-autoregressive Machine Translation}\ 
\rrrevise{\citet{nat2018gu} propose NAT models to reduce the latency in generation or decoding, but there exists a gap in translation quality between NAT and AT models.}
To bridge the gap, iterative NATs manage to repeatedly refine the generated outputs \cite{iterativerefinement2018lee, cmlm2019ghazvininejad, jmnat2020guo}. However, as shown in \citet{deepshallow2021kasai}, most iterative NATs are not advantageous against ATs in the quality-latency tradeoff.
Non-iterative NATs are much faster, whose improvements mainly come from alignment-based objectives \cite{ctc2018libovicky, axe2020ghazvininejad, oaxe2021du}, or incorporating extra decoder inputs \cite{vae2020shu, glat2021qian, CNAT2021bao}.
Nevertheless, these NATs heavily rely on knowledge distillation (KD, \citealp{nat2018gu}), which is found very effective in reducing the data modalities \cite{kdnat2020zhou}.
A recent study \cite{mple2021} provides a unified perspective showing that most existing methods actually modify targets or inputs to reduce the token dependencies in the data distribution, which eases the NAT training but introduces data distortion.

Unlike existing NATs, our method retains multiple translations instead of dropping the multi-modal information in NAT training. It turns out that our method can effectively tackle the multi-modality problem without modifying the training data and not rely on KD to achieve a good translation performance.

\begin{figure*}[!t]
    \centering
    \includegraphics[width=0.75\linewidth]{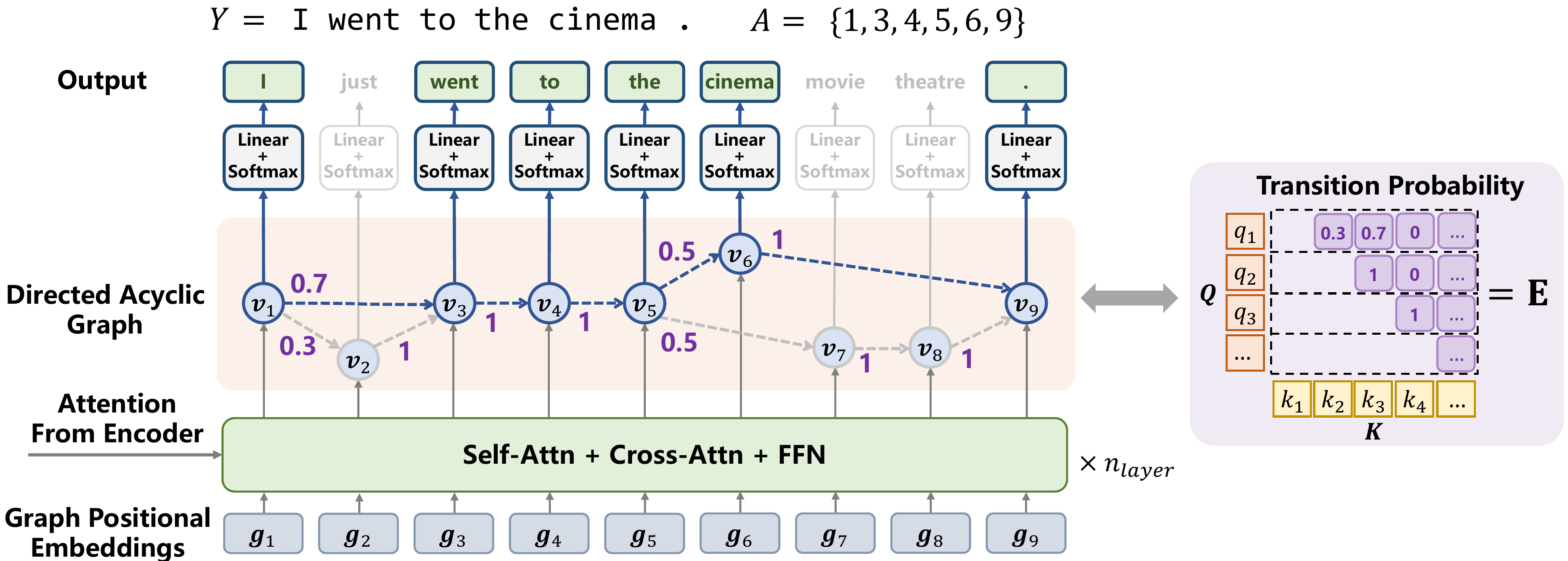}
    \vspace{-0.5em}
    \caption{Overview of Directed Acyclic Decoder. The decoder organizes the hidden states $\mathbf{V}=[\bv_1, \cdots, \bv_L]^T$ in a directed acyclic graph (DAG) structure, which stores representations of multiple translations. 
    The \textcolor{mypurple}{transition probabilities in $\mathbf{E}$} are predicted based on the vertex states.
    To generate one of the translations, a path $A$ is first sampled from the DAG structure, and the \textcolor{myblue}{selected vertices} predict the whole sentence in parallel. The \textcolor{mygray}{other vertices} and their predicted tokens are skipped in the output.}
    \label{fig:model}
    \vspace{-1em}
\end{figure*}

\textbf{Lattice-based Model in Machine Translation}\ \ \ \ 
\textit{Word lattices} have a long history in Statistic Machine Translation. A word lattice is a directed acyclic graph (DAG) with edges labeled with a token and weight, which can represent an exponential number of sentences in the a compact structure.
A phrase-based translation system can generate a word lattice during decoding \cite{DBLP:conf/emnlp/UeffingON02, DBLP:journals/coling/OchN04}. 
Some models take word lattices as inputs to alleviate input errors brought by word segmentation or speech recognition \cite{DBLP:conf/acl/DyerMR08, DBLP:conf/acl/KoehnHBCFBCSMZDBCH07, DBLP:conf/coling/DongCLXSIH14}.
There are also studies that combine multiple system outputs into a single lattice \cite{DBLP:conf/naacl/RostiAXMSD07, DBLP:conf/emnlp/FengLMLL09} and decode a good translation from it \cite{DBLP:conf/emnlp/TrombleKOM08}.

\rrrevise{Unlike previous studies that construct the word lattices with a search algorithm, our model predicts the whole DAG with multiple translations simultaneously. Moreover, \methodname's training does not require ground-truth word lattices for supervision, making it applicable to most translation benchmarks.}

\section{Our Proposed Method}

In this section, we describe our proposed \methodname in detail.
Intuitively, to facilitate the explicitly modeling of multiple modalities, we propose to replace the original Non-autoregressive Transformer decoder with a directed acyclic decoder, whose topological structure is a DAG.
Each path of the \graphname forms a sequence of hidden states that stores a possible translation, and the whole \graphname store multiple translations in different paths.
\methodname still generates in a non-autoregressive fashion.

We will first introduce the network structures in Section \ref{sec:model}, which presents how to construct the \methodname to parameterize the conditional probability.
Then in Section \ref{sec:training}, we will elaborate on the training of \methodname, including how to train it with one reference and the efficient implementation of traversing possible paths.
Finally, in Section \ref{sec:inference}, we provide several decoding approaches, aiming to sample fluent sentences efficiently given the well-trained \methodname.

\subsection{Architecture of \methodname}
\label{sec:model}

\methodname consists of a Transformer encoder and a \textit{directed acyclic decoder}.
The encoder is the same as vanilla Transformer while the decoder organizes its hidden states as a DAG.
As shown in Fig.\ref{fig:model}, hidden states correspond to vertices of the DAG, which model word distributions in specific positions; and edges of the DAG are transitions between hidden states, which organize generated words into a final sentence.

Intuitively, given a source sentence $X$, the directed acyclic decoder generates a sentence in three steps:
(1) receiving the position embeddings as inputs and producing hidden states as vertices;
(2) calculating the transition probabilities between the vertices based on the vertex states;
(3) sampling a path from the \graphname following the transitions, and then predicting target tokens using the vertex states on the path.

Formally, the probability of a target sentence $Y = \{ y_1, y_2, \cdots, y_M \}$ is formulated as
\begin{align}
    P_{\theta}(Y|X) &= \sum_{A \in \Gamma} P_{\theta}(Y, A|X) \notag \\
    &= \sum_{A \in \Gamma} P_{\theta}(A|X) P_{\theta}(Y|A, X), \label{eq:main}
\end{align}
where $A=\{ a_1, a_2, \cdots, a_M \}$ is a path represented by a sequence of vertex indexes, and $\Gamma$ contains all paths with the same length of the target sentence $Y$. %namely $\{1 = a_1 \leq \cdots \leq a_M = L \}$.

\textbf{Vertex}\ \ \ 
The directed acyclic decoder utilizes the Transformer layers~\cite{transformer2017vaswani} to predict the vertex states.
Unlike the autoregressive decoder that generates tokens from left to right, it generates the vertex states in parallel.

\rrrevise{Specifically, we use graph positional embeddings $\mathbf{G} = \{ \bg_1, \cdots, \bg_L \}$ as the decoder inputs,
which is identical to the learnable positional embeddings in vanilla Transformer but represents the vertex indexes instead of the token positions.}
Note that $L$ is the graph size, where we set $L$ to $\lambda$ times the source length $N$ and tune $\lambda$ as a hyper-parameter.
The decoder then produces the vertex states $\mathbf{V}=[\bv_1, \cdots, \bv_L]^T$, which is defined as
\begin{equation}
    [\bv_1, \cdots, \bv_L] = \text{Transformer-Blocks}(\bg_1, \cdots, \bg_L). \notag
\end{equation}
\textbf{Transition}\ \ \ \ 
Each edge of the \graphname is assigned the transition probability between the connecting vertices. The transition probabilities are locally normalized, i.e., the probabilities of outgoing edges sum to one.
Formally, the probability of path $A$ is defined as
\begin{gather}
    P_{\theta}(A|X) = \prod_{i=1}^{M-1} P_{\theta}(a_{i+1}|a_{i}, X) =  \prod_{i=1}^{M-1} \mathbf{E}_{a_{i}, a_{i+1}}, \notag
\end{gather}
where $\mathbf{E} \in \mathbb{R}^{L \times L}$ is the transition matrix normalized by rows.
Specifically, the transition matrix is obtained by
\begin{gather}
    \mathbf{E} = \text{softmax}(\frac{\mathbf{Q} \mathbf{K}^T}{\sqrt{d}}), \label{eq:matrixE} \\
    \mathbf{Q} = \mathbf{V}  \mathbf{W}_\text{Q},\quad \mathbf{K} = \mathbf{V} \mathbf{W}_\text{K},  \notag
\end{gather}
where $d$ is the hidden size, $\mathbf{W}_\text{Q}$ and $\mathbf{W}_\text{K}$ are learnable parameters.
To ensure that there is no cycle in the \graphname, we apply lower triangular masking on $\mathbf{E}$, which only allows transitions from vertices with small indexes to large indexes.
Note that the matrix $\mathbf{E}$ can be calculated in parallel, thereby facilitating fast sampling of paths.

\textbf{Token Prediction}\ \ \ \ 
Conditioned on the vertex states in $\mathbf{V}$ and the selected path $A$, the decoder predicts the target tokens in parallel. Formally, we have
\begin{align}
P_{\theta}(Y|A, X) = \prod_{i=1}^{M} P_{\theta}(y_i|a_i, X) = \prod_{i=1}^{M} \text{softmax}(\mathbf{W}_\text{P} \bv_{a_i}), \notag
\end{align}
where $\mathbf{W}_\text{P}$ are learnable weights, \rrrevise{and $\bv_{a_i}$ is the representation of the $i$-th vertex on the path $A$.}

In the implementation, we actually calculate the distributions on all vertices and then skip the vertices not appearing on the chosen path.
Specifically, we obtain
\begin{align}
    \mathbf{P} = \text{softmax}(\mathbf{V} \mathbf{W}_\text{P}^T ), \label{eq:matrixP}
\end{align}
where $\mathbf{P} \in \mathbb{R}^{L \times |\mathbb{V}|}$ is the matrix containing the token distributions on the $L$ vertices, and $P_{\theta}(y_i|a_i, X) = \mathbf{P}_{a_i, y_i}$.
The matrix $\mathbf{P}$ facilitates fast calculation for multiple paths since the shared vertices are not calculated twice, which is significant in training and inference introduced later.

\subsection{Training}
\label{sec:training}

To capture multiple translation modalities in training, the proposed directed acyclic decoder arranges words from different modalities in different vertex states of the decoder, which can effectively reduces the inconsistent problem in training.
In this section, we will elaborate on training details of \methodname, including training with one reference, efficient implementation of marginalizing all paths in the DAG, and modified glancing training techniques according to the graph structures.

\rrrevise{\textbf{Training \methodname with One Reference}}\ \ \ \ 
Although \methodname retains multiple translations in the \graphname, its training objective only requires one reference per sample, which facilitates efficient training on most translation benchmarks.
Specifically, it directly maximizes the log-likelihood $\log P(Y|X)$ by marginalizing all possible paths $A$, which can be formulated as follows,
\begin{align}
    \Lmain = - \log P_{\theta}(Y|X) = - \log \sum_{A \in \Gamma} P_{\theta}(Y, A|X), \label{eq:L_main}
\end{align}
where $\Gamma$ contains all paths with $1 = a_1 < \cdots < a_M = L$.

\rrrevise{To understand why a single reference is adequate for the DAG learning, we analyze the training process by inspecting the gradients.}
\revise{Intuitively, we find that the objective assigns a single reference to several paths, where the vertices on the chosen paths are updated to generate the reference tokens, and the other vertices remain unchanged. The sparse assignment is the key to the successful training, which avoids inconsistent labels in token predictions and preserves the unseen translations stored on the unchanged paths.} \rrrevise{In such a way, the DAG can be learned across different training instances, each of which only provides a single reference, not requiring an instance with multiple references.}

\revise{Specifically, we inspect the gradient of $\Lmain$ and find that}
\begin{align}
    \frac{\partial}{\partial \theta} \Lmain = \sum_{A \in \Gamma} w_A \left( \frac{\partial}{\partial \theta} \LpathA \right),  \label{eq:gradient_analysis}
\end{align}
where
\begin{align}
    \LpathA &= - \log P_{\theta}(Y, A|X),  \\
    w_A &= \frac{P_{\theta}(Y, A|X)}{\sum_{A' \in \Gamma} P_{\theta}(Y, A'|X)}.  \label{eq:gradient_weight}
\end{align}
$\LpathA$ maximizes the likelihood of sampling $Y$ with the path $A$, and $w_A$ is the weight of $\LpathA$. Eq(\ref{eq:gradient_analysis}) indicates that the weights of paths are assigned according to the probability that $Y$ appears on $A$. 
If a path $A$ is more probable for the target $Y$, then a larger weight will be used in optimizing $\LpathA$, which further strengthens its dominance. In contrast, an unlikely path $A$ will get a negligible weight, indicating that the vertices on $A$ are not affected in the update.

\begin{figure}[!t]
    \centering
    \includegraphics[width=\linewidth]{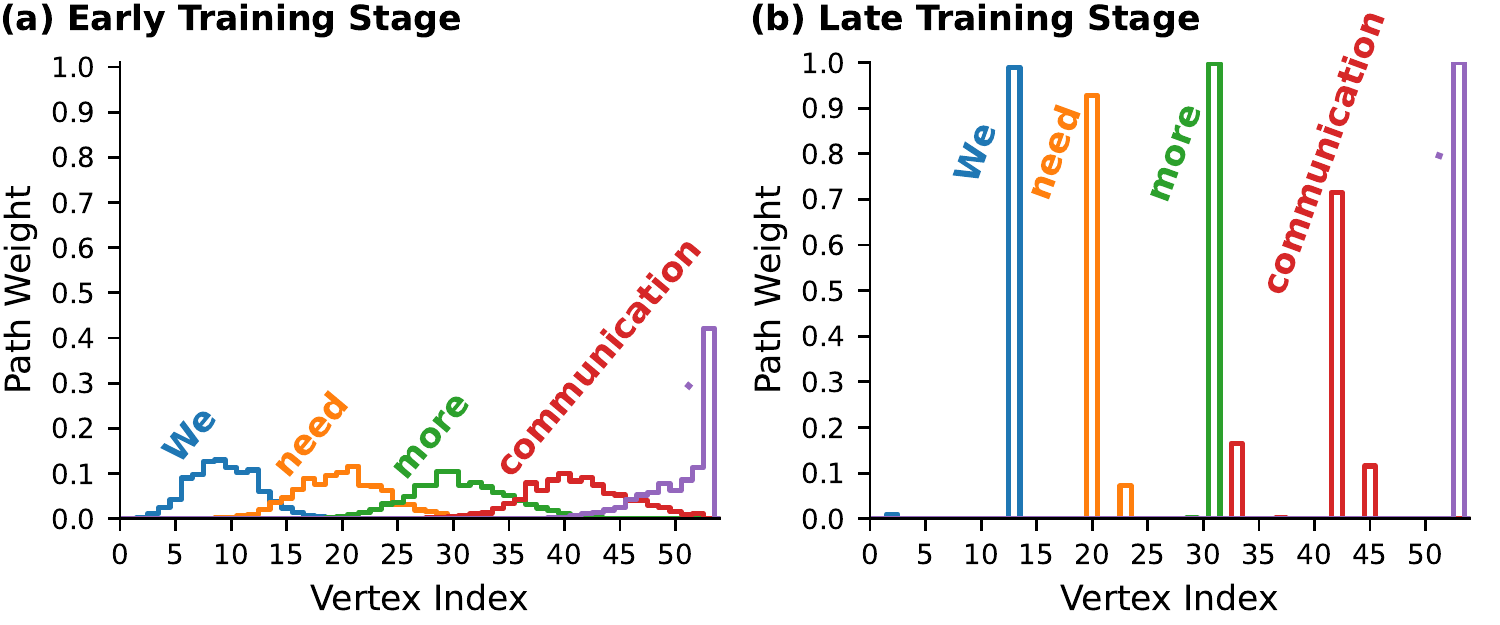}
    \vspace{-1.8em}
    \caption{The cumulative weight $w_A$ of paths that pass through each vertex, where lines represent token labels for the vertex.\protect\footnotemark~The weights are sparse in the late stage of training, indicating that only several vertices are updated to fit the sample. Source: ``\textit{我们 需要 更多 的 交流 。}''  Target: ``\textit{We need more communication .}''}
    \label{fig:converge}
    \vspace{-0.5em}
\end{figure}

A real example is shown in Fig.\ref{fig:converge}. In the early stage, training with one sample will affect all vertices in the \graphname. In the late stage, only some vertices are updated, reserving the other vertices for storing unseen translations.

\addtocounter{footnote}{-1}

\footnotetext{For example, the orange line (the second token, \textit{need}) on the vertex $v_{20}$ is the sum of $w_A$ for the paths $A=\{a_1, a_2, \cdots, a_5\}$ satisfying $a_2 = 20$. $w_A$ is defined in Eq(\ref{eq:gradient_weight}).}

\textbf{Marginalizing $A$ with Dynamic Programming} \ 
The objective $\Lmain$ requires margin\-al\-izing all paths $A$, which is expensive due to the numerous paths.
Similar to \citet{ctc2006graves}, we employ dynamic programming to tackle the issue.

\revise{Generally, we recurrently calculate the probability sum of path prefixes that end at the vertex $u$ and generate the target prefix $Y_{\leq i}$, denoted as $f_{i, u}$. Since the path prefixes that end at the vertex $u$ should pass through a vertex $v$ satisfying $v < u$, so $f_{i, u}$ can be obtained from $f_{i-1, v}$.}
\rrevise{By recurrently calculating $f_{i, u}$, we finally obtain the probability sum of all valid paths and the training objective $\Lmain = - \log f_{M, L}$.}
\revise{Our algorithm reduces the time complexity to $\mathcal{O}(ML^2)$ and can be implemented by $\mathcal{O}(M)$ PyTorch operations. The detailed formulation is presented in Appendix \ref{app:dp}.}

\begin{figure}[!t]
    %\vspace{-1em}
    \centering
    \includegraphics[width=0.9\linewidth]{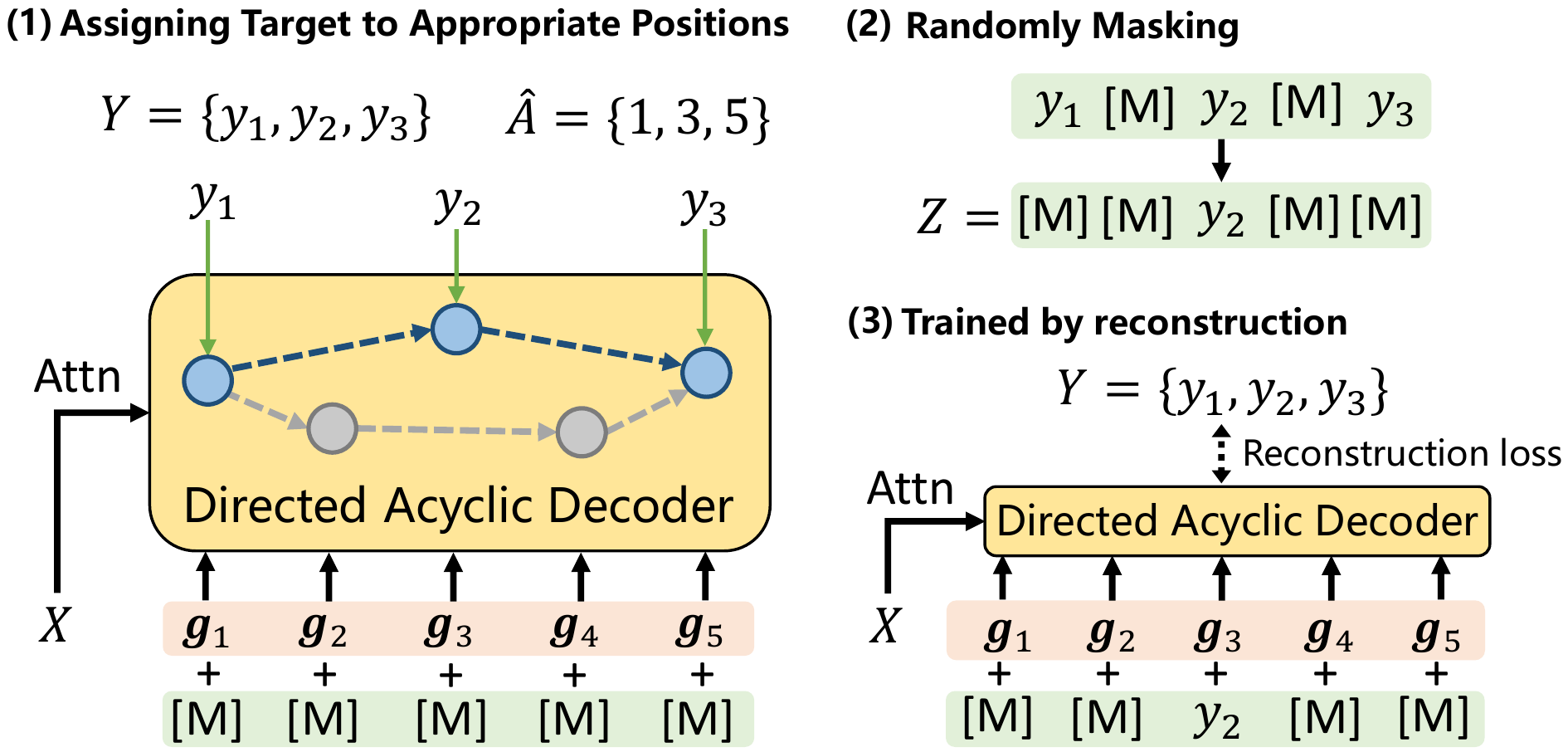}
    \vspace{-0.5em}
    \caption{Glancing training for \methodname, which is similar to the masked language model for promoting representation learning. Glancing training requires two forward passes of Directed Acyclic Decoder: The first pass assigns target tokens to appropriate positions. The second pass calculates the reconstruction loss. \textsf{[M]} indicates a masked token whose embeddings are all zeros.}
    \label{fig:mask-predict}
    \vspace{-1em}
\end{figure}

\textbf{Glancing Training on Graph}\ \ \ \ 
Previous work shows that glancing training \cite{glat2021qian} can significantly improve the translation quality of non-iterative NATs.
Here we present a modified glancing training technique on DAG, \revise{which still requires only one reference per sample.}

Specifically, to improve the training of \methodname via glancing training, we add a masked target to the decoder input and train the model by reconstruction, which promotes the learning of dependency between vertices.
Formally, the objective of glancing is defined as
\begin{align}
    \Lmp & = - \log P_{\theta}(Y|X, Z), \label{eq:Lmp}
\end{align}
where $Z=[z_1, \cdots, z_L]$ is a randomly masked target provided as an extra decoder input, and $P_{\theta}(Y|X, Z)$ is similarly defined as Eq(\ref{eq:L_main}). 

The glancing training follows three steps, as shown in Fig.\ref{fig:mask-predict}.
(1) We assign the target tokens to appropriate vertices since the decoder input is longer than the target sentence.
The assignment follows the most probable path $\hat{A} = \argmax_{A \in \Gamma} P_\theta(Y, A|X)$, which requires a forward pass of the decoder and dynamic programming. 
(2) We obtain $Z$ by masking some tokens. We utilize the masking strategy proposed by GLAT \cite{glat2021qian}, which decides the number of unmasked tokens according to the prediction accuracy.\footnote{The number of unmasked token $t = \tau \sum_{i=1}^{M} [y_i \neq \hat{y}_i]$, where $\hat{y}_i = \argmax P_\theta(\cdot|a_i, X)$, and $\tau \in [0, 1]$ is a hyper-parameter.}
(3) We add $Z$ to the decoder input and train the model by minimizing Eq(\ref{eq:Lmp}).

\subsection{Inference}
\label{sec:inference}

In inference, \methodname constructs a \graphname that stores multiple translations, where we aim to find the most probable one.
Compared with existing NATs, \methodname  utilizes transitions to distinguish different candidates, which improves fluency and avoids errors like repeated tokens.
We propose three decoding strategies to find high-quality translations while keeping low latency.

\addtolength{\textfloatsep}{-2em} 
\begin{algorithm}[!t]
   \caption{\footnotesize Greedy / Lookahead Decoding in Pytorch-like 
    \\ \text{\quad\quad\quad\quad\quad\ \ \ } Parallel Pseudocode}
   \label{algo:greedy}
\begin{algorithmic}
\begin{footnotesize}
   \STATE {\bfseries Input:} Graph Size $L$, Transition Matrix $\mathbf{E} \in \R^{L \times L}$, \\
   \quad\quad\quad Token Distributions $\mathbf{P} \in \R^{L \times |\mathbb{V}|}$ 
   \IF{Using Lookahead}
   \STATE $\mathbf{E} := \mathbf{E} \otimes [\mathbf{P}$.\textsc{max}({dim}=1).\textsc{unsqueeze}({dim}=0)$]$
   \STATE \# $\mathbf{E}$ now jointly considers $\mathbf{P}$ and $\mathbf{E}$
   \STATE \# $\otimes$ is element-wise multiplication
   \ENDIF
   \STATE $\textit{tokens} := \mathbf{P}$.\textsc{argmax}({dim}=1) \quad \text{\# shape: (L)}
   \STATE $\textit{edges} := \mathbf{E}$.\textsc{argmax}({dim}=1) \quad \text{\# shape: (L)}
   \STATE $i := 1$, $\textit{output} :=$ [ \textit{tokens}[$1$] ]
   \REPEAT
   \STATE $i := $ \textit{edges}[$i$] \quad \text{\# jumping along the transition}
   \STATE \textit{output}.\textsc{append}(\textit{tokens}[$i$])
   \UNTIL{$i = L$}
   \end{footnotesize}
\end{algorithmic}
\end{algorithm}

\textbf{Greedy}\ \ \ \ 
The simplest strategy is to take the most likely choices for the transitions and tokens.
Specifically, we perform parallel argmax operations to obtain the most likely transition and token for each vertex.
Then, we generate the translation by collecting the predicted tokens along the chosen path. 
The greedy decoding is highly efficient that only uses two parallel operations, as shown in Algo.\ref{algo:greedy}.

\addtolength{\textfloatsep}{2em} 

\textbf{Lookahead}\ \ \ \ 
Lookahead decoding improves the greedy strategy by jointly considering the transitions and the tokens.
Specifically, we rearrange $P_{\theta}(Y, A|X)$ into
\begin{align}
     P_{\theta}(y_1|a_1, X) \prod_{i=2}^{M} P_{\theta}(a_{i}|a_{i-1}, X) P_{\theta}(y_{i}|a_{i}, X), \label{eq:sequence_decode}
\end{align}
which becomes a sequential decision problem of choosing $a_{i}$ and $y_{i}$ in order. We simultaneously obtain
\begin{align}
y^*_{i}, a^*_{i} = \argmax P_{\theta}(y_{i}|a_{i}, X) P_{\theta}(a_{i}|a_{i-1}, X), \label{eq:joint}
\end{align}
which can be still implemented in parallel with almost zero overhead, as presented in Algo.\ref{algo:greedy}.

\begin{table*}[t]
\caption{Results on WMT14 En$\leftrightarrow$De and WMT17 Zh$\leftrightarrow$En. \rrrrevise{We present \methodname' results with mean and standard deviation of three runs with different random seeds. Best performance of non-iterative NATs (iter=1) are \textbf{bolded}. \textsf{Average Gap} is the gap of BLEU against the best AT model, excluding the missing values.}
* indicates results of our re-implementation. Our autoregressive transformer is better than previously reported results because we use the same training setting as NATs (300k steps, 64k tokens/batch; previous results use 100k steps, 32k tokens/batch). $^\dag$ uses reranking methods in NAT decoding (LPD, \citealp{imitatenat2019wei}). }
\label{tab:main_result}
\vspace{-0.5em}
\begin{center}
\begin{small}
%\begin{sc}
\resizebox{\linewidth}{!}{
\setlength{\tabcolsep}{1mm}{
\begin{tabular}{l|c|>{\hspace*{1mm}}r@{}lr@{}l|>{\hspace*{1mm}}r@{}lr@{}l|>{\hspace*{1mm}}r@{}lr@{}l|>{\hspace*{1mm}}r@{}lr@{}l|>{\hspace*{0.5mm}}r@{}l>{\hspace*{2.5mm}}r@{}l|>{\hspace*{2mm}}r@{}l}
\toprule
\multirow{2}{*}{\bf Model} & \bf Iter & \multicolumn{4}{c|}{\bf WMT14 En-De} & \multicolumn{4}{c|}{\bf  WMT14 De-En} & \multicolumn{4}{c|}{\bf WMT17 En-Zh} & \multicolumn{4}{c|}{\bf WMT17 Zh-En} & \multicolumn{4}{c|}{\bf Average Gap $\downarrow$} & \multicolumn{2}{c}{\multirow{2}{*}{\bf  Speedup}}\\
& $\#$ & \multicolumn{2}{c}{Raw} & \multicolumn{2}{c|}{KD} & \multicolumn{2}{c}{Raw} & \multicolumn{2}{c|}{KD} & \multicolumn{2}{c}{Raw} & \multicolumn{2}{c|}{KD} & \multicolumn{2}{c}{Raw} & \multicolumn{2}{c|}{KD} &  \multicolumn{2}{c}{\hspace*{2mm}Raw} & \multicolumn{2}{c|}{KD} & \\
\midrule
Transformer~{\scriptsize \cite{transformer2017vaswani}} & $M$ & 27&.6 & 27&.8 & 31&.4 & 31&.3 & 34&.3 & 34&.4 & 23&.7 & 24&.0 & \tempneg0&.45 & \tempneg0&.49 & 1&.0x \\
Transformer (Ours) & $M$ & 28&.07* & 28&.54* & 31&.94* & 31&.54* & 34&.89* & 34&.69* & 23&.89* & 24&.68* & 0& & 0& & 1&.0x \\
\midrule
CMLM~{\scriptsize \cite{cmlm2019ghazvininejad}} & 10 & 24&.61 & 27&.03 & 29&.40 & 30&.53 & \multicolumn{2}{c}{-} & 33&.19 & \multicolumn{2}{c}{-}  & 23&.21 & \tempneg3&.00 & \tempneg1&.37 & 2&.2x\\
SMART~{\scriptsize \cite{smart2020}} & 10 & 25&.10 & 27&.65 & 29&.58 & 31&.27 & \multicolumn{2}{c}{-} & 34&.06 & \multicolumn{2}{c}{-} & 23&.78 & \tempneg2&.67 & \tempneg0&.67 & 2&.2x\\
DisCo~{\scriptsize\cite{disco2020kasai}} & $\approx$4 & 25&.64 & 27&.34 & \multicolumn{2}{c}{-} & 31&.31 & \multicolumn{2}{c}{-} & 34&.63 & \multicolumn{2}{c}{-} & 23&.83 & \tempneg2&.43 & \tempneg0&.59 & 3&.5x \\
Imputer~{\scriptsize\cite{imputermt2020saharia}} & 8 & 25&.0 & 28&.2 & \multicolumn{2}{c}{-} & 31&.8 & \multicolumn{2}{c}{-} & \multicolumn{2}{c|}{-} & \multicolumn{2}{c}{-} & \multicolumn{2}{c|}{-} & \tempneg3&.07 & \tempneg0&.04 & 2&.7x \\
CMLMC~{\scriptsize\cite{cmlmc2021}} & 10 & 26&.40 & 28&.37 & 30&.92 & 31&.41 & \multicolumn{2}{c}{-} & \multicolumn{2}{c|}{-} & \multicolumn{2}{c}{-} & \multicolumn{2}{c|}{-} & \tempneg1&.35 & \tempneg0&.15 & 1&.7x \\
\midrule
Vanilla NAT~{\scriptsize\cite{nat2018gu}} & 1 & 11&.79* & 19&.99* & 16&.27* & 25&.77* & 18&.92* & 25&.84* & 8&.69* & 14&.81* & \tempneg15&.78 & \tempneg8&.26 & 15&.3x \\
CTC~{\scriptsize\cite{ctc2018libovicky}} & 1 & 18&.42* & 25&.52 & 23&.65* & 28&.73 & 26&.84* & 31&.39* & 12&.23* & 19&.93* & 9&.41 & 3&.47 & 14&.6x \\
AXE$^\dag$~{\scriptsize\cite{axe2020ghazvininejad}} & 1 & 20&.40 & 23&.53 & 24&.90 & 27&.90 & \multicolumn{2}{c}{-} & 30&.88 & \multicolumn{2}{c}{-} & 19&.79 & \tempneg7&.36 & \tempneg4&.34 & 14&.2x \\
GLAT~{\scriptsize\cite{glat2021qian}} & 1 & 19&.42* & 25&.21 & 26&.51* & 29&.84 & 29&.79* & 32&.22* & 18&.88* & 21&.84* & \tempneg6&.05 & \tempneg2&.59 & 15&.3x \\
OaXE$^\dag$~{\scriptsize\cite{oaxe2021du}} & 1 & 22&.4 & 26&.1 & 26&.8 & 30&.2 & \multicolumn{2}{c}{-} & 32&.9 & \multicolumn{2}{c}{-} & 22&.1 & \tempneg5&.4 & \tempneg2&.0 & 14&.2x \\
CTC + GLAT~{\scriptsize\cite{glat2021qian}} & 1 & 25&.02* & 26&.39 & 29&.14* & 29&.54 & 30&.65* & 32&.51* & 19&.92* & 23&.11* & \tempneg3&.52 & \tempneg1&.98 & 14&.6x \\
CTC + DSLP~{\scriptsize\cite{dslp2021huang}} & 1 & 24&.81 & 27&.02 & 28&.33 & 31&.61 & \multicolumn{2}{c}{-} & \multicolumn{2}{c|}{-} & \multicolumn{2}{c}{-} & \multicolumn{2}{c|}{-} & \tempneg3&.44 & \tempneg0&.73 & 14&.0x \\
\midrule
\methodname + Greedy (Ours)\ \  & 1 & 26&.08{\stdfont{$\pm.25$}} & 27&.31{\stdfont{$\pm.08$}} & 30&.48{\stdfont{$\pm.18$}} & 31&.30{\stdfont{$\pm.06$}} & 33&.27{\stdfont{$\pm.12$}} & 33&.80{\stdfont{$\pm.11$}} & 22&.66{\stdfont{$\pm.12$}} & 24&.04{\stdfont{$\pm.09$}} & \tempneg1&.58 & \tempneg0&.75 & 14&.0x\\
+ Lookahead & 1 & 26&.57{\stdfont{$\pm.21$}} & 27&.49{\stdfont{$\pm.05$}} & 30&.68{\stdfont{$\pm.24$}} & 31&.37{\stdfont{$\pm.06$}} & 33&.83{\stdfont{$\pm.13$}} & 34&.08{\stdfont{$\pm.13$}} & 22&.82{\stdfont{$\pm.20$}} & 24&.23{\stdfont{$\pm.14$}} & \tempneg1&.22 & \tempneg0&.57 & 13&.9x \\
+ BeamSearch & 1 & 27&.02{\stdfont{$\pm.15$}} & 27&.78{\stdfont{$\pm.07$}} & 31&.24{\stdfont{$\pm.18$}} & 31&.80{\stdfont{$\pm.03$}} & 34&.21{\stdfont{$\pm.21$}} & \bf 34&\bf.35{\stdfont{$\pm.12$}} & 24&.22{\stdfont{$\pm.10$}} & 24&.90{\stdfont{$\pm.16$}} & \tempneg0&.53 & \tempneg0&.16 & 7&.1x\\
+ BeamSearch + $5$-gram LM & 1 & \bf 27&.25{\stdfont{$\pm.12$}} & \bf 27&.91{\stdfont{$\pm.07$}} & \bf 31&\bf.54{\stdfont{$\pm.20$}} & \bf 31&.95{\stdfont{$\pm.06$}} & \bf 34&\bf.23{\stdfont{$\pm.17$}} & 34&.27{\stdfont{$\pm.05$}} & \bf 24&\bf.49{\stdfont{$\pm.06$}} & \bf 25&\bf.01{\stdfont{$\pm.18$}} & \bf \tempneg0&\bf.32 & \bf \tempneg0&\bf.08 &  7&.0x \\
\bottomrule
\end{tabular}
}}
%\end{sc}
\end{small}
\end{center}
\vspace{-1em}
\end{table*}

\textbf{BeamSearch}\ \ \ \ 
BeamSearch is a more accurate method for solving the above decoding problem. 
%Inspired by \citet{ctcbeamsearch2014hannun}
Following \citet{tricktrade2021gu}, we combine an $n$-gram language model to improve the performance. Specifically, we search in beam to approximately find the optimal $Y^*$ that maximizes
\begin{align}
\frac{1}{|Y|^{\alpha}} \left[ \log P_{\theta}(Y|X) + \gamma \log P_{n\text{-gram}}(Y) \right] , \label{eq:beamsearch}
\end{align}
where $\alpha, \gamma$ are hyper-parameters for length penalty and language model scores.
\rrrevise{Note that $Y$ can appear on multiple paths, where we obtain the probability sum of these paths in BeamSearch to obtain $P_{\theta}(Y|X)$.
More details are shown in Appendix \ref{app:beamsearch}.}

It should be noticed that BeamSearch requires sequential operations and does not preserve the non-autoregressive nature. However, such sequential operations do not involve deep network computations and can still be very efficient, with
about 7 times speedups compared with AT models.\footnote{We will release an efficient C++ implementation at \url{https://github.com/thu-coai/DA-Transformer}.}

\section{Experiments}

\begin{figure*}[!t]
\begin{minipage}[p]{.6\textwidth}
\centering
\includegraphics[width=0.9\textwidth]{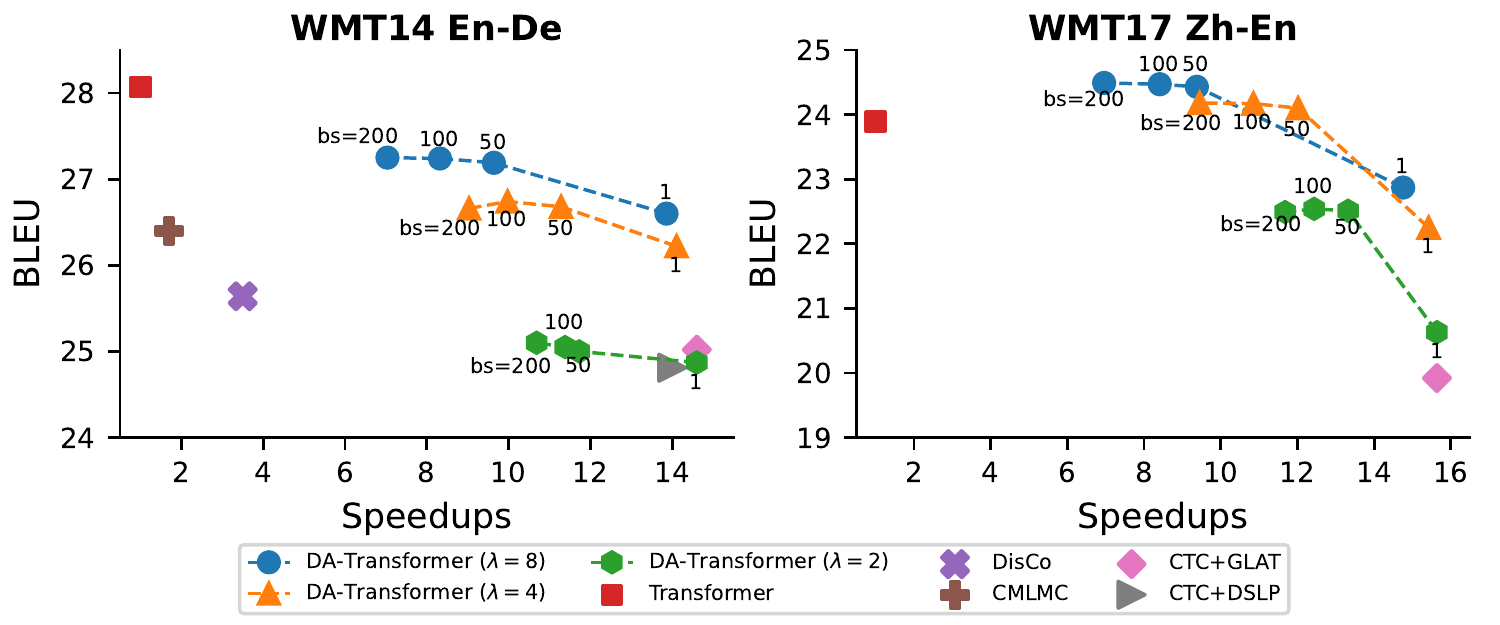}
\vspace{-0.5em}
\caption{\revise{Quality-latency tradeoff on WMT14 En-De and WMT17 Zh-En with varying graph sizes and beam sizes. The graph size is $\lambda$ times of the source length. \rrrrevise{We use Beamsearch + 5-gram LM with the beam size (bs) of $200, 100, 50$.} $\text{bs}=1$ indicates Lookahead Decoding.}}
\label{fig:tradeoff}
\end{minipage}
\hfill
\begin{minipage}[p]{.36\textwidth}
\centering
%\vspace{-0.5em}
\includegraphics[width=0.82\textwidth]{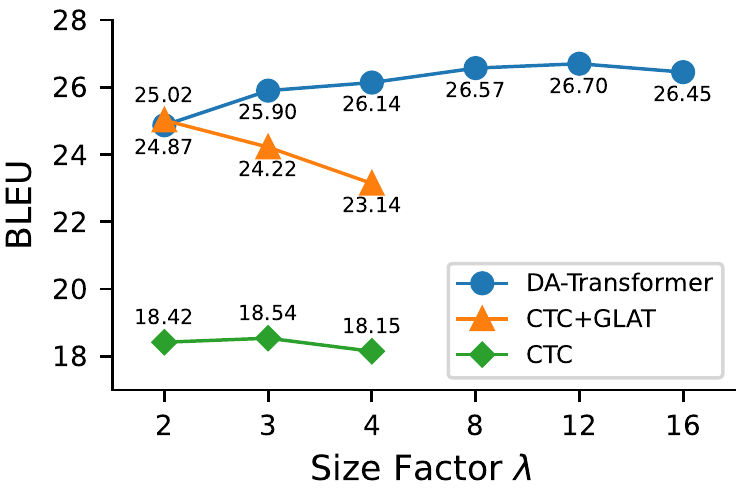}
\vspace{-0em}
\caption{Effects of $\lambda$ on WMT14 En-De. The graph size (\methodname) or the output length (CTC) is $\lambda$ times of the source length. \label{fig:upsample}}
\end{minipage}
\vspace{-0.8em}
\end{figure*}

\noindent \textbf{Dataset}\ \  
We conduct experiments on two benchmarks, WMT14 En$\leftrightarrow$De (4.5M) and WMT17 Zh$\leftrightarrow$En (20M), where we follow \citet{kdnat2020zhou, disco2020kasai} for pre-processing.
For knowledge distillation, we follow \citet{oaxe2021du} to use Transformer-\textit{big} as our teacher model and generate the distilled data with a beam size of 5.

\noindent \textbf{Metrics}\ \  
For fair comparisons with previous work, we use tokenized BLEU \cite{bleu2002papineni} for all benchmarks except WMT17 En-Zh, where we use sacreBLEU \cite{sacrebleu2018}. The latency speedup is evaluated on WMT17 En-De test set with a batch size of 1.

\noindent \textbf{Hyper-parameters}\ \ 
Our models generally use the hyper-parameters of transformer-\textit{base} \cite{transformer2017vaswani}. For regularization, we set dropout to 0.1, weight decay to 0.01, and label smoothing to 0.1.
All models, including ATs, are trained for 300k updates with a batch of 64k tokens. The learning rate warms up to $5 \cdot 10^{-4}$ within 10k steps and then decays with the inverse square-root schedule.
We evaluate the BLEU scores on the validation set every epoch and average the best 5 checkpoints for the final model.
For \methodname, we use $\lambda=8$ and Lookahead Decoding unless otherwise specified. 
We linearly anneal $\tau$ from 0.5 to 0.1 for glancing training. 
For BeamSearch, we set beam size to 200, $\gamma$ to 0.1, and tune $\alpha$ from $[1, 1.4]$ on the validation set.
The training lasts approximately 32 hours on 16 Nvidia V100-32G GPUs.

\subsection{Main Results}

As shown in Table \ref{tab:main_result},
\methodname substantially improves the translation quality and outperforms strong baselines by a large margin.
Our model alleviates the multi-modality problem by capturing multiple translations within a DAG, which avoids inconsistent labels in training and reduces the errors of mixing translations in inference.
We highlight the empirical advantages of our method:

1) Better translation quality compared with non-iterative NATs.
As a non-iterative NAT, our model achieves new SoTA results in translation quality while preserving competitive speedups.
Unlike existing NATs which heavily rely on KD, our model with Lookahead outperforms the best baselines on the raw data by \rrrrevise{2.2 BLEU on average}, verifying that \methodname can effectively alleviate the multi-modality problem without simplifying the training data.

2) Lower inference latency compared with ATs and iterative NATs. 
\methodname achieves 7x$\sim$14x speedups over ATs, where the remaining BLEU gaps are about \rrrrevise{0.32} on average. Especially on WMT17 Zh-En, our best model with BeamSearch outperforms ATs by 0.6 BLEU.
Moreover, \methodname dominates all iterative NATs on both BLEU and latency for all benchmarks except WMT14 En-DE with KD, which shows the great potential of our model.

3) Flexible quality-latency tradeoff. 
Comparing the decoding strategies of our method, we find that Lookahead Decoding consistently outperforms Greedy Decoding, and the n-gram LM \rrrrevise{usually} benefits BeamSearch, with almost zero overheads.
\rrrrevise{To better show the quality-latency tradeoff, we tune the graph size and beam size with our decoding strategies.}
As shown in Fig.\ref{fig:tradeoff}, our method significantly outperforms existing NATs and provides flexible quality-latency tradeoff for non-autoregressive translation.

\subsection{Ablation Study}

In this section, we investigate the effects of the graph size and training methods on the raw data of WMT14 En-De.

\textbf{Graph Size}\ \ 
\methodname utilizes a \graphname with $L$ vertices, which is empirically set to $\lambda$ times of the source length. A large \graphname can model more translations. However, it also makes the transition predictions difficult. We manually tune $\lambda$ from 2 to 16, as shown in Fig.\ref{fig:upsample}.

The results show that larger graphs improve the translation quality until $\lambda$ exceeds 12, where $\lambda$ is not sensitive around its best value. 
We compare our methods against CTC, which also utilizes a similar hyper-parameter to determine the output length \cite{ctc2018libovicky}. Although CTC+GLAT has a similar performance with \methodname when $\lambda=2$, the BLEU score does not increase for a larger $\lambda$.
We attribute the problem to the inconsistent label problem: a longer output sequence does not help CTC to reduce the inconsistent labels in training, where \methodname benefits from larger graphs by assigning different tokens to distinct vertices.
Considering the performance and computation cost, we choose $\lambda=8$ and apply it to all other datasets.

\begin{figure}[!t]
    \centering
    \includegraphics[width=0.75\linewidth]{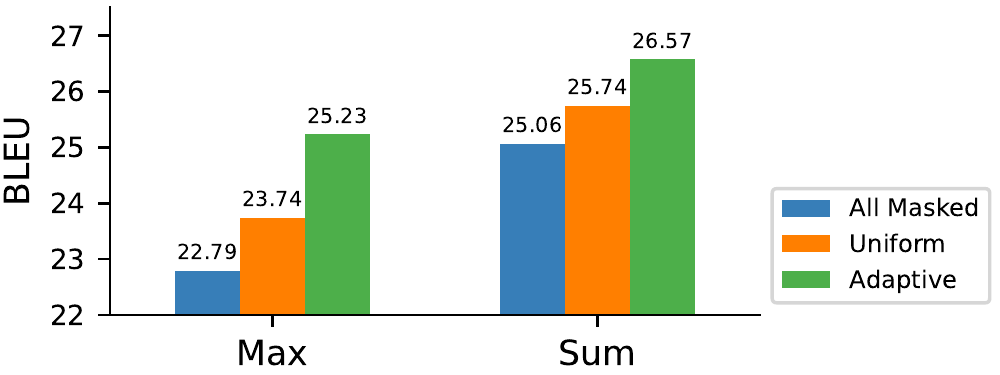}
    \caption{Ablation study of training objectives on WMT14 En-De. \textbf{Max}/\textbf{Sum} represent using the max operation or the sum operation in marginalizing all paths in Eq(\ref{eq:L_main}). We compare three masking strategies in obtaining the decoder input: \textbf{All Masked}, \textbf{Uniform} \cite{cmlm2019ghazvininejad}, and \textbf{Adaptive} \cite{glat2021qian}.}
    \label{fig:mlm}
    \vspace{-1.5em}
\end{figure}

\textbf{Training Objectives}\ \ \ \ 
\methodname is trained with a glancing objective that only requires one reference for each sample, where we investigate two important designs:
\textit{First}, we marginalize all possible paths to obtain $\Lmain$, which is equivalent to optimizing the paths with different weights as discussed in Sec.\ref{sec:training}. We compare it with the objective only optimizing the most probable path, i.e., replacing the sum operation by the max operation in Eq(\ref{eq:L_main}).
\textit{Second}, we use glancing training with a masked target as inputs, where the masked tokens are adaptively chosen according to the prediction accuracy \cite{glat2021qian}.
We compare it with two other strategies: masking all inputs (i.e., do not use glancing training), uniform random masking \cite{cmlm2019ghazvininejad}.

The results are shown in Fig.\ref{fig:mlm}.
\textit{First}, marginalizing all paths (Sum) outperforms choosing the most probable path (Max). One possible reason is that the max operation makes sharp weight assignments in the early training, leading to a premature convergence in which only several paths are used.
\textit{Second}, the glancing training (Uniform or Adaptive) is better than the vanilla training (All Masked), which improves the translation quality by promoting representation learning.
Moreover, the adaptive strategy can further boost performance by choosing the masking ratio dynamically.

\begin{table}[!t]
\caption{Token accuracy under \textit{the best assignment}. An assignment matches each reference token with a predicted token, which is called an alignment in CTC or a path in \methodname.}
\label{tab:acc}
\vskip 0.1in
\begin{center}
\begin{small}
%\begin{sc}
\begin{tabular}{l|cc|cc}
\toprule
\multirow{2}{*}{\bf Model} & \multicolumn{2}{c|}{\bf WMT14 En-De} & \multicolumn{2}{c}{\bf WMT17 Zh-En} \\
& Train & Valid & Train & Valid \\
\midrule
Vanilla NAT & 29.7 & 29.6 & 39.8 & 22.4  \\
CTC + GLAT & 50.2 & 51.7 & 47.1 & 32.3 \\
\midrule
\methodname & \bf 69.3 & \bf 69.9 & \bf 80.1 & \bf 67.0 \\
\bottomrule
\end{tabular}
%\end{sc}
\end{small}
\end{center}
\vspace{-1em}
\end{table}

\subsection{Analysis}

This section verifies that \methodname benefits from assigning tokens to vertices in training and explicitly considers the transitions in inference. It also shows some cases of learned DAGs. \rrrevise{We present more analyses in the appendix, including the translation performance on different lengths (Appendix \ref{app:length_bucket}), performance with controlled training time (Appendix \ref{app:traincurve}), and some statistics of DAGs (Appendix \ref{app:dag_stats}).}

\textbf{\methodname improves token accuracy.}\ \ 
In training, we assign tokens of different translations to different vertices, which avoids the inconsistent labels in training and thus improves the token accuracy in inference.
We compare our model against two baselines, Vanilla NAT and CTC+GLAT. Note that CTC utilizes an alignment-based objective, which also assigns the reference tokens to different positions of Transformer. 
We calculate the accuracy \textit{under the best assignment} following two steps: We first obtain the most probable assignment that matches each reference token to a prediction. Then, we calculate the accuracy by comparing the predicted tokens on the best assignment (i.e., the best path in \methodname) against the reference. \footnote{In CTC, a reference token may be matched with several predictions, so we average the accuracies for the reference token. The special empty tokens are not counted in the accuracy.}

As shown in Table \ref{tab:acc}, vanilla NAT suffers from label inconsistency problem, leading to low token accuracies. Comparing \methodname and CTC, we find that our method is far more effective, especially on the syntax distant language pair such as WMT17 Zh-En.
We conjecture that the advantage mainly comes from our flexible assignment method. CTC only avoids position mismatches by inserting empty or repeated tokens.
It requires that the possible translations share similar lexical choices, which cannot handle highly diverse translations.

\begin{figure}[!t]
    \centering
    \includegraphics[width=\linewidth]{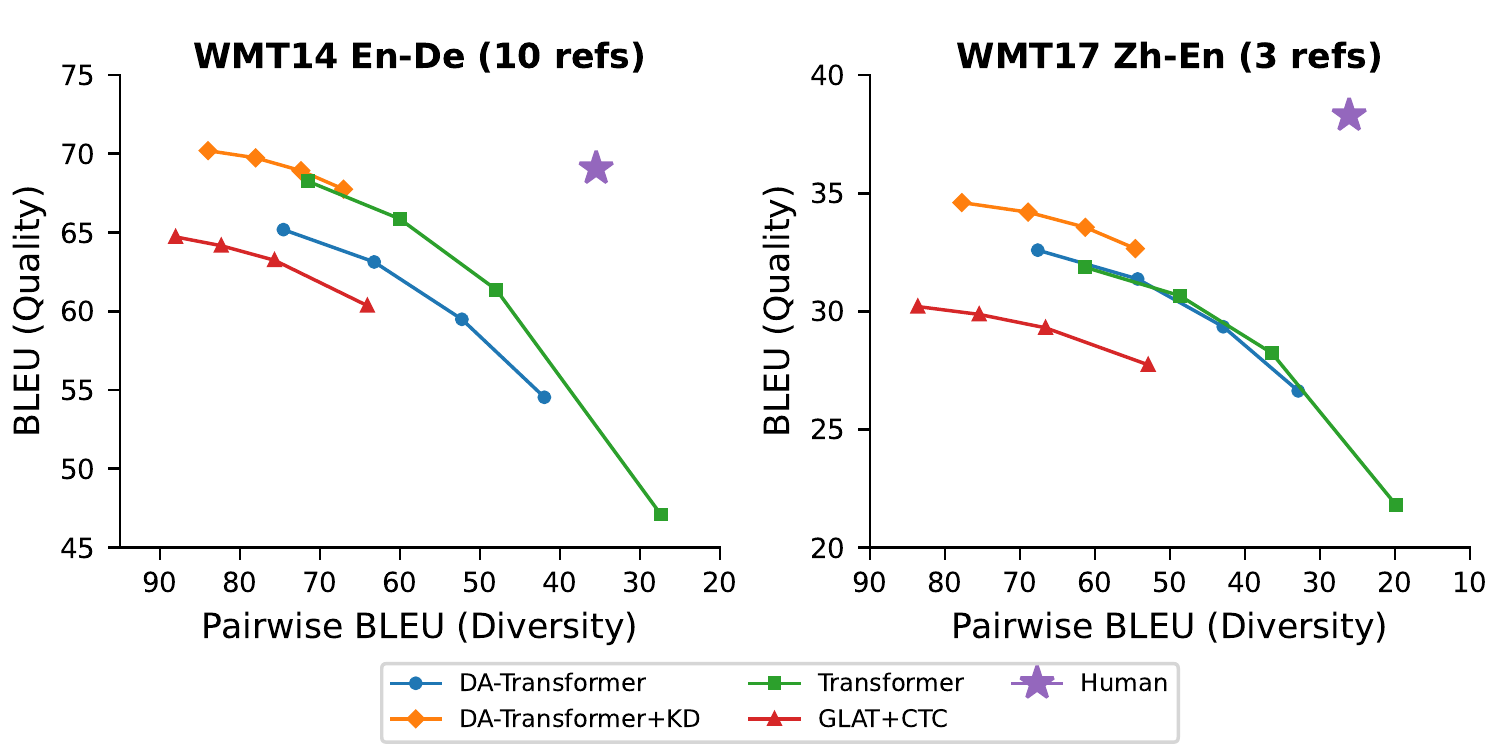}
    \vspace{-2em}
    \caption{\revise{Quality-diversity tradeoff of sampled sentences from different models. Each curve has 4 points with sampling temperatures $t=[0.4, 0.6, 0.8, 1]$ (from left to right). We sample hypotheses with the same number of human-written references.}}
    \label{fig:diversity}
    \vspace{-1.5em}
\end{figure}

\begin{figure*}[!t]
    \centering
    \includegraphics[width=0.95\linewidth]{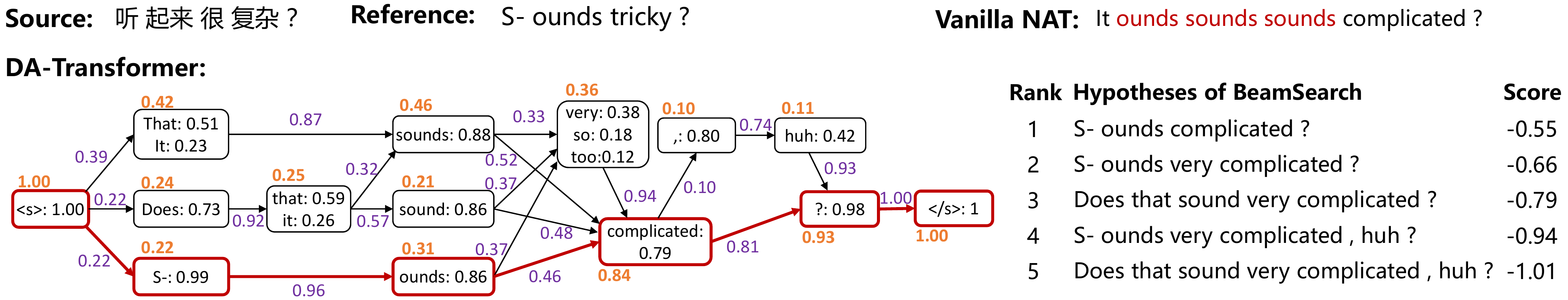}
    \caption{A test sample of WMT17 Zh-En with the \graphname and BeamSearch results.
    We present the \textsf{top candidates} on each vertex and the \textcolor{mypurple}{transition probabilities} between vertices. We also present the \textcolor{myorange}{passing probabilities} representing how likely a vertex will appear on a sampled path. We remove vertices/edges with small passing/transition probabilities for a clear presentation.
    The vanilla NAT mixes the tokens from different translations, which can be avoided in \methodname inference.
    We use the BPE tokenizer \cite{bpe2016sennrich}, where a subword prefix is marked by \textsf{-}. 
    See more examples in Appendix \ref{app:more_cases}.
    }
    \label{fig:case}
    \vspace{-1em}
\end{figure*}

\textbf{\methodname facilitates diverse generation.}\ \ \ \ 
In inference, \methodname utilizes the transition matrix to avoid incorrect outputs caused by mixing multiple translations.
We evaluate the ability to distinguish different translations by sampling diverse translations from the \graphname.
\revise{Specifically, we begin at the start vertex and repeatedly use Nucleus Sampling (top-p sampling, \citealp{topp2020holtzman}) to choose the next vertex and token according to Eq.(\ref{eq:sequence_decode}).}
We use $p=0.8$ and vary the temperature from 0.4 to 1.0.

We follow \citet{mixturemodel2019shen} to evaluate the quality and diversity by multi-reference BLEU and pairwise BLEU.
We compare our model against AT and GLAT+CTC, the best non-iterative NAT baseline. We obtain the hypotheses from GLAT+CTC by replacing the argmax operations in decoding with Nucleus Sampling with the same $p$ and temperature.

The results are shown in Fig.\ref{fig:diversity}.
Compared with GLAT+CTC, \methodname achieves a better tradeoff between quality and diversity.
With the same temperature, the generated samples (without KD) by our model are far more diverse than GLAT+CTC. It shows that our model can learn multiple diverse translations and further decode them in inference.
Compared with Transformer, \methodname (without KD) is slightly less diverse but achieves a close tradeoff on WMT17 Zh-En, which shows the great potential of our model.
\revise{Moreover, we find that applying KD to \methodname improves the quality but sacrifices the diversity because KD reduces the data modalities.}

\textbf{Case Study}\ \ \ \ 
We choose a test sample of WMT17 Zh-En and illustrate the \graphname predicted by our model.
For a clear presentation, we use $\lambda=4$ for a small graph and further remove some useless vertices and edges.
\revise{Specifically, we remove all vertices with passing probabilities smaller than 0.1, where the passing probabilities represent how likely the vertex will appear on a randomly sampled path. We only show the transitions in the top 90\% of probabilities.}

As shown in Fig.\ref{fig:case}, the predicted \graphname is highly reasonable.
Following the transitions, we can clearly distinguish translation expressions, which avoids the errors like repeated tokens shown in the vanilla NAT's output.
We present the top-5 hypotheses produced by BeamSearch, which are fluent and diverse.

\revise{However, we can still find errors in the predicted \graphname, e.g., a possible incorrect translation ``\textit{Does that sounds ...}''. Although the error does not easily occur in Lookahead or BeamSearch decoding, it shows that there is still space for improving the consistency between the tokens in our model.}

\section{Conclusion}
In this paper, we propose \methodname for non-autore\-gressive machine translation. 
Unlike previous NAT models relying on knowledge distillation, \methodname tackles the multi-modality problem by capturing \rrevise{multiple translations} with a directed acyclic decoder.
Experimental results show that \methodname outperforms all NAT baselines on raw training data and achieves competitive results with AT models.
The best model of \methodname even outperforms the autoregressive Transformer by 0.6 BLEU on Zh-En, which demonstrates the potential of the proposed approach.

\nocite{discretelatent2018kaiser}
\nocite{enhancedecoderinput2019guo}
\nocite{crf2019sun}
\nocite{pnat2019bao}
\nocite{bagofngram2020shao}
\nocite{emnatsun2020}
\nocite{continuousrefine2020}
\nocite{posconstrain2021yang}
\nocite{lowfreq2021ding}
\nocite{fairseq2019ott}

\section*{Acknowledgments}

This work was supported by the National Science Foundation for Distinguished Young Scholars (with No. 62125604) and the NSFC projects (Key project with No. 61936010 and regular project with No. 61876096). This work was also supported by the Guoqiang Institute of Tsinghua University, with Grant No. 2019GQG1 and 2020GQG0005, and sponsored by Tsinghua-Toyota Joint Research Fund.

\bibliography{cite}
\bibliographystyle{icml2022}

%%%%%%%%%%%%%%%%%%%%%%%%%%%%%%%%%%%%%%%%%%%%%%%%%%%%%%%%%%%%%%%%%%%%%%%%%%%%%%%
%%%%%%%%%%%%%%%%%%%%%%%%%%%%%%%%%%%%%%%%%%%%%%%%%%%%%%%%%%%%%%%%%%%%%%%%%%%%%%%
% APPENDIX
%%%%%%%%%%%%%%%%%%%%%%%%%%%%%%%%%%%%%%%%%%%%%%%%%%%%%%%%%%%%%%%%%%%%%%%%%%%%%%%
%%%%%%%%%%%%%%%%%%%%%%%%%%%%%%%%%%%%%%%%%%%%%%%%%%%%%%%%%%%%%%%%%%%%%%%%%%%%%%%
\newpage
\appendix
\onecolumn

\section{\revise{Dynamic Programming for Training}}
\label{app:dp}

\begin{figure}[!h]
    \centering
    \includegraphics[width=0.85\linewidth]{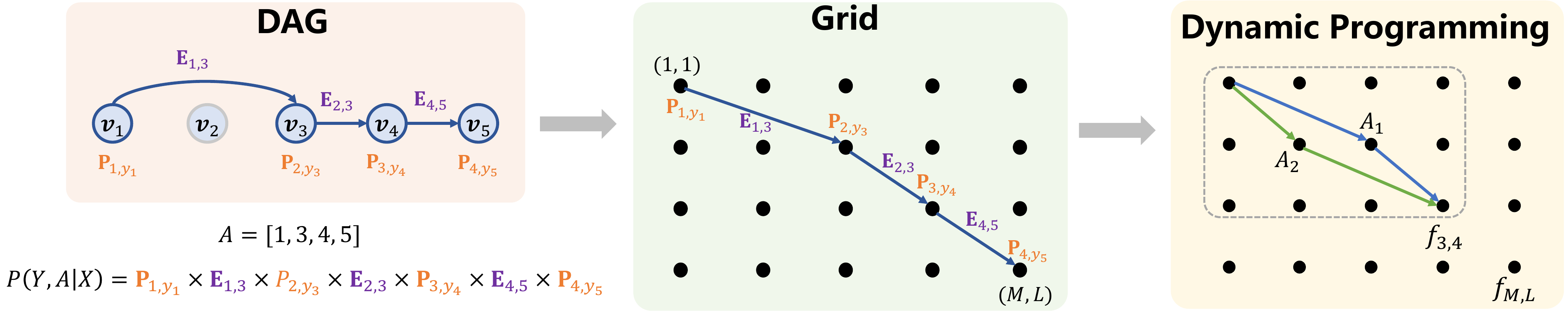}
    \vspace{-0.3em}
    \caption{An overview of dynamic programming for DA-Transformer training. Each valid path in the DAG corresponds to a path in the grid that starts from $(1, 1)$ to $(M, L)$. $M$ is the target length, $L$ is the graph size. $P(Y, A|X)$ can be calculated by multiplying \textcolor{myorange}{the token probabilities} and \textcolor{mypurple}{the transition probabilities}. In Dynamic Programming, we recurrently calculate $f_{i, u}$, which is the probability sum of all paths that start from $(1, 1)$ and end at $(i, u)$. E.g., $f_{3, 4}$ is the sum of two paths' probabilities, $A_1$ and $A_2$. Our training objective $\Lmain$ is equal to $-\log f_{M, L}$.}
    \label{fig:appendix_dp}
    \vspace{-0em}
\end{figure}

\begin{algorithm}[!b]
    \small
   \caption{Dynamic Programming Algorithm in Pytorch-like Parallel Pseudocode}
   \label{algo:dp}
\begin{algorithmic}
   % \small
   \STATE {\bfseries Input:} Target Length $M$, Graph Size $L$, Target Sentence $Y$, Transition Matrix $\mathbf{E} \in \R^{L \times L}$, Token Distributions $\mathbf{P} \in \R^{L \times |\mathbb{V}|}$
   \STATE Initialize a zero matrix $f \in \R^{M \times L}$
   \STATE $f[1, 1] := 1$
   \FOR{$i = 2, 3, \cdots, M$}
   \STATE $f[i, :] := \mathbf{P}[:, y_i] \otimes (f[i-1, :] \times \mathbf{E})$ \quad \text{\# $\otimes$ is the element-wise multiplication, $\times$ is the vector-matrix multiplication}
   \ENDFOR
   %\STATE $\text{tokens} := \mathbf{P}$.argmax(dim=1) \quad \text{\# shape: (L)}
   \STATE Update the model by minimizing $\Lmain = - \log f[M, L]$.
\end{algorithmic}
\end{algorithm}

The training objective of DA-Transformer is formulated in Eq(\ref{eq:L_main}), which requires marginalizing all possible paths $A$. To avoid the expensive cost of enumerating the paths, we employ dynamic programming that reduces the time complexity to $\mathcal{O}(ML^2)$, where $M$ is the target length, $L$ is the graph size.

To utilize dynamic programming, we first represent the valid paths in a $M \times L$ grid, where each valid path of the DAG corresponds  to a path in the grid that starts from the left-upper corner $(1, 1)$ and ends in the right-bottom corner $(M, L)$, as shown in Fig.\ref{fig:appendix_dp}.
Formally, the path $A=\{ a_1, \cdots, a_M \}$ satisfying $1 = a_1 < \cdots < a_M = L$ corresponds to a path in the grid that passes through $(1, a_1), (2, a_2), \cdots, (M, a_M)$.

Then we find that the probability on the path $A$, i.e., $P_{\theta}(Y, A|X)$, can be decomposed and calculated by multiplying the probabilities of the token predictions and transitions.
Specifically, we have
\begin{align}
P_{\theta}(Y, A|X) &= \prod_{j=1}^{M} P_{\theta}(y_{j}|a_{j}, X) \prod_{j=2}^{M} P_{\theta}(a_{j}|a_{j-1}, X)  \\
    &= \prod_{j=1}^{M} \mathbf{P}_{a_{j}, y_{j}} \prod_{j=2}^{M} \mathbf{E}_{a_{j-1}, a_{j}}.
\end{align}
where $\mathbf{P}_{a_{j}, y_{j}}$ can be regarded as the token probability on the point $(j, a_{j})$, and the $\mathbf{E}_{a_{j-1}, a_{j}}$ can be regarded as the transition probability on the edge connecting $(j-1, a_{j-1})$ with $(j, a_j)$.

Recall that our objective requires the sum of the probabilities of all valid paths.
We can recurrently calculate $f_{i, u}$, which is defined as the probability sum of the paths that start from $(1,1)$ but end at $(i, u)$. %Our objective $\Lmain = - \log f_{M, L}$. 
Since each valid path that ends at $(i, u)$ must pass through a point $(i-1, v)$ where $v \in [1, u)$, we reach a recurrence formula that obtains $f_{i, u}$ from $f_{i-1, v}$:
\begin{gather}
    f_{i, u} = \mathbf{P}_{u, y_i} \sum_{v=1}^{u-1} f_{i-1, v} \  \mathbf{E}_{v, u} \quad (2 \leq i \leq M, 1 \leq u \leq L),
\end{gather}
where the boundary conditions are:
\begin{gather}
    f_{1, 1} = \mathbf{P}_{1, y_1};\quad\quad\quad\quad f_{1, u} = 0 \quad (2 \leq u \leq L).
\end{gather}

Finally, the loss can be obtained by $\Lmain = -\log f_{M, L}$. 

Since the product-sum operations can be calculated by matrix multiplications, the above recurrent process can be implemented with $\mathcal{O}(M)$ parallel operations, as shown in Algorithm \ref{algo:dp}.

\section{\revise{Implementation of Beam Search}}
\label{app:beamsearch}

\begin{figure}[!h]
    \centering
    \vspace{1em}
    \includegraphics[width=\linewidth]{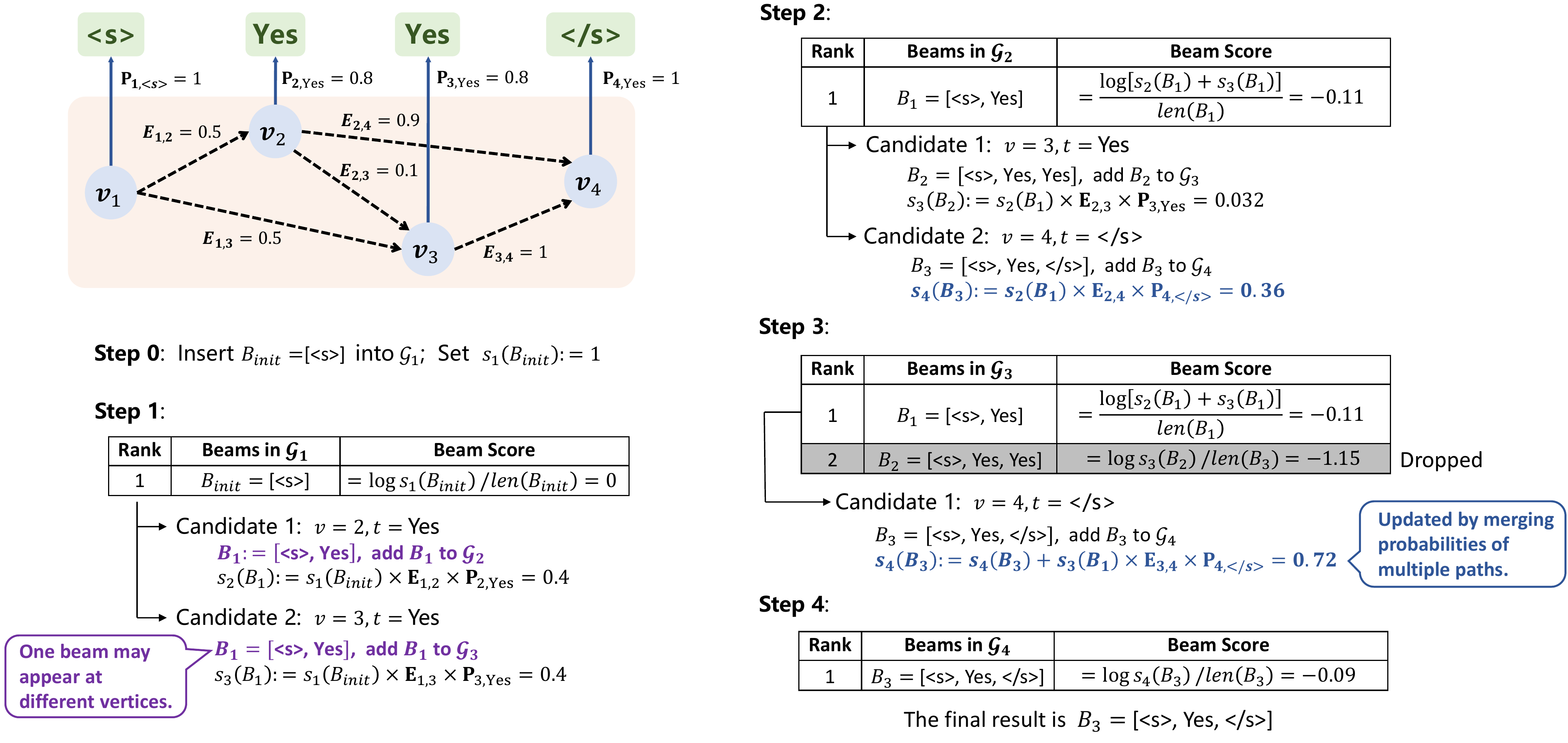}
    \vspace{-1.5em}
    \caption{An step-by-step example of the beam search algorithm presented in Algorithm \ref{algo:beamsearch}. One beam represents a translation prefix, which may appear on multiple paths. For demonstration, we only preserve the top-1 beam in each step, limit the candidate number when expanding the beams, and set $\alpha=1$ for the length penalty and $\gamma=0$ to disable the n-gram language model.}
    \label{fig:appendix_beamsearch}
    \vspace{0em}
\end{figure}

%\addtolength{\textfloatsep}{2em}

Our concept of beam is similar to the prefix beam search \cite{ctcbeamsearch2014hannun}, where a beam represents a translation prefix but may appear on multiple paths.
E.g., in Fig.\ref{fig:appendix_beamsearch}, $[\langle \text{s} \rangle, \text{Yes}]$ is a beam that appears on two paths, $\{v_1, v_2\}$ and $\{v_1, v_3\}$.
Our beam search aims to calculate the probability sum of all paths that produce the same translation, which approximates $P(Y|X)$ and works better than finding a single path that maximizes $P(Y, A|X)$.

To achieve an effective calculation of the scores, we maintain the probability sum for a beam $B$ during the beam search. Specifically, we define $s_i(B)$ as the probability sum of the paths ends at vertex $i$. When sorting the beams, we use the beam score defined in Eq(\ref{eq:beamsearch}), where $P(Y|X)$ is equal to the probability sum of all paths, i.e., $\sum_{i=1}^{L} s_i(B)$.

Our algorithm is presented in Algorithm \ref{algo:beamsearch} with an example shown in Fig.\ref{fig:appendix_beamsearch}.
We further apply some tricks to reduce the computation costs:
\vspace{-0.5em}

\begin{itemize}[leftmargin=1em]
    \setlength{\itemsep}{0ex}{
    \setlength{\parskip}{2px}{
    \item Unlike vanilla beam search that all beams have the same length in each step, our algorithm may compare beams with different lengths. To avoid a length bias in the selected beams, we only preserve the top-$10$ for each length. If the total number of beams is still too large, we choose the top-$200$ beams.
    \item When expanding beams, we only use the top-$5$ candidates. A candidate is a $\langle v, t \rangle$ pair, indicating the next vertex and token, where we jointly consider their probabilities as Eq(\ref{eq:joint}).
    }
    }
\end{itemize}
\vspace{-0.5em}

\begin{algorithm}[!t]
    \small
   \caption{BeamSearch for \methodname}
   \label{algo:beamsearch}
\begin{algorithmic}
   \STATE {\bfseries Input:} Graph Size $L$, Transition Matrix $\mathbf{E} \in \R^{L \times L}$, Token Distributions $\mathbf{P} \in \R^{L \times |\mathbb{V}|}$
   %, Transition Matrix $\mathbf{E} \in \R^{L \times L}$, Token Distributions $\mathbf{P} \in \R^{L \times |\mathbb{V}|}$
   \STATE For $i \in [1, L]$ and any possible beam $B$, initialize $s_i(B):=0$ that stores the probability sum of all $B$'s paths that end at vertex $i$
   \STATE Initialize $\mathcal{G}_1, \cdots, \mathcal{G}_L$ as $L$ empty sets that store the beams at the vertex $i$
   \STATE Insert a beam $B_{init}$ with the start token in $\mathcal{G}_1$. Set $s_1(B_{init}) := 1$
   \FOR{$i = 1, 2, \cdots, L-1$}
   \STATE \textcolor{mygreen}{\# Filter the beams}
   \STATE Sort the beams in $\mathcal{G}_i$ by the score defined in Eq(\ref{eq:beamsearch})
   \STATE For each beam length in $\mathcal{G}_i$, preserve the top-10 beams and remove the other beams
   \STATE Considering all beams in $\mathcal{G}_i$, preserve the top-200 beams and remove the other beams
   \STATE \textcolor{mygreen}{\# Expand the beams}
   \FOR{each beam $B$ in $\mathcal{G}_i$}
    \FOR{$\langle v, t \rangle$ in $B$'s top-5 next candidates}
    \STATE \textcolor{mygreen}{\# $v$ is the next vertex; $t$ is the next token}
    \STATE Insert $B' = B + \{t\}$ as a new beam into $\mathcal{G}_v$.
    \STATE Update $s_v(B') := s_v(B') + s_i(B) \times \mathbf{P}_{v,t} \times \mathbf{E}_{i, v}$
    \ENDFOR
   \ENDFOR
   \ENDFOR
   \STATE Output the best beam in $\mathcal{G}_L$
\end{algorithmic}
\end{algorithm}

\section{\revise{More Analyses}}

\subsection{\revise{Translation Performance on Different Lengths}}

\label{app:length_bucket}

\begin{figure}[!h]
    \centering
    \vspace{-1em}
    \includegraphics[width=0.5\linewidth]{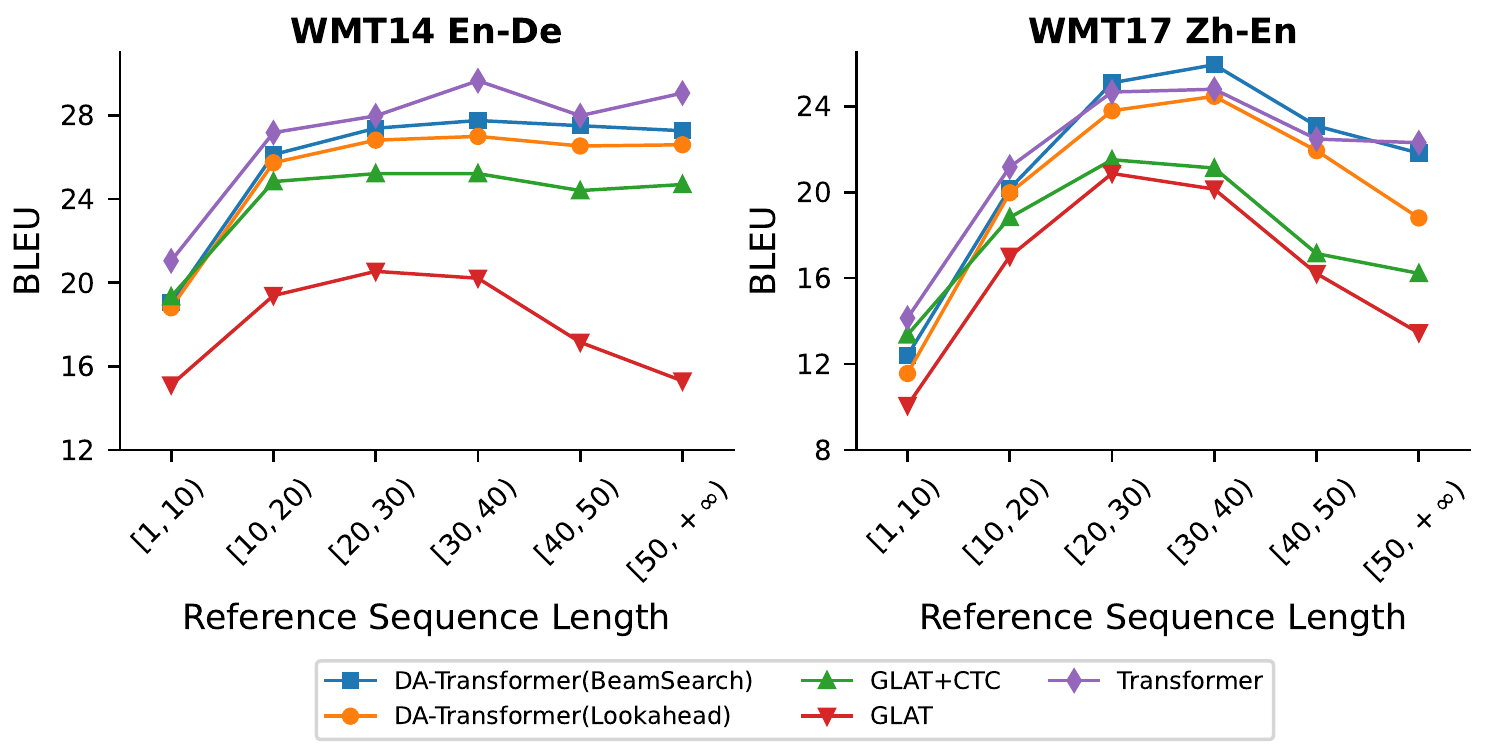}
    \vspace{-0.5em}
    \caption{The BLEU score on WMT14 En-De and WMT17 Zh-En bucketed by the reference length.}
    \label{fig:lengthbleu}
    \vspace{-1em}
\end{figure}

To investigate the translation performance on different lengths, we split the test set into 6 buckets according to reference
lengths and evaluate the BLEU score in each bucket as shown in Fig.\ref{fig:lengthbleu}.
Compared with NAT baselines, \methodname has a substantial improvement for sentences longer than 20. These long sentences usually have more modalities in translation, which are challenging in previous NATs but can be better handled in \methodname.

\subsection{\revise{Performance with Controlled Training Time}}

\label{app:traincurve}

\begin{figure}[!h]
    \vspace{-1em}
    \centering
    \includegraphics[width=0.4\linewidth]{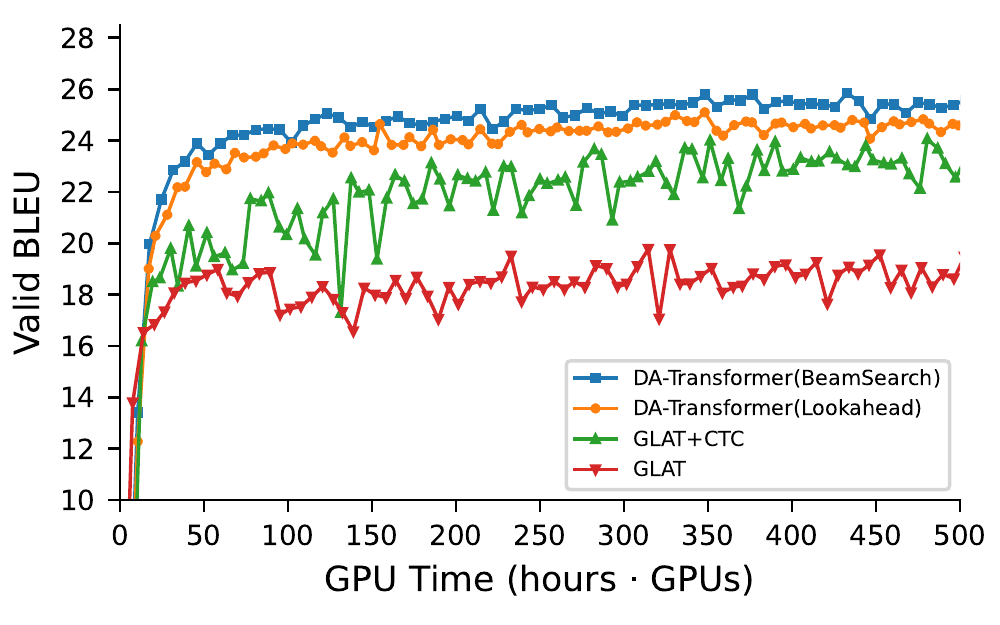}
    \vspace{-1em}
    \caption{The valid BLEU on WMT14 En-De. We do not apply the checkpoint average trick. We evaluate the model approximately every 8 GPU-hours. The training costs 500 GPU-hours, which has about 300k, 490k, 970k updates for DA-Transformer, GLAT+CTC, GLAT, respectively.}
    \label{fig:traincurve}
    \vspace{0em}
\end{figure}

One update step of \methodname's training is slower than many previous NATs because our Directed Acyclic Decoder has to process a longer sequence whose length is about 8 times of the original target. In Fig.\ref{fig:traincurve}, we show that \methodname still substantially outperforms strong NAT baselines when the training time is controlled. Moreover, we observe that our performance is more stable than the baselines during the training process.

\section{More Cases}

\begin{figure}[!h]
    \centering
    \includegraphics[width=\linewidth]{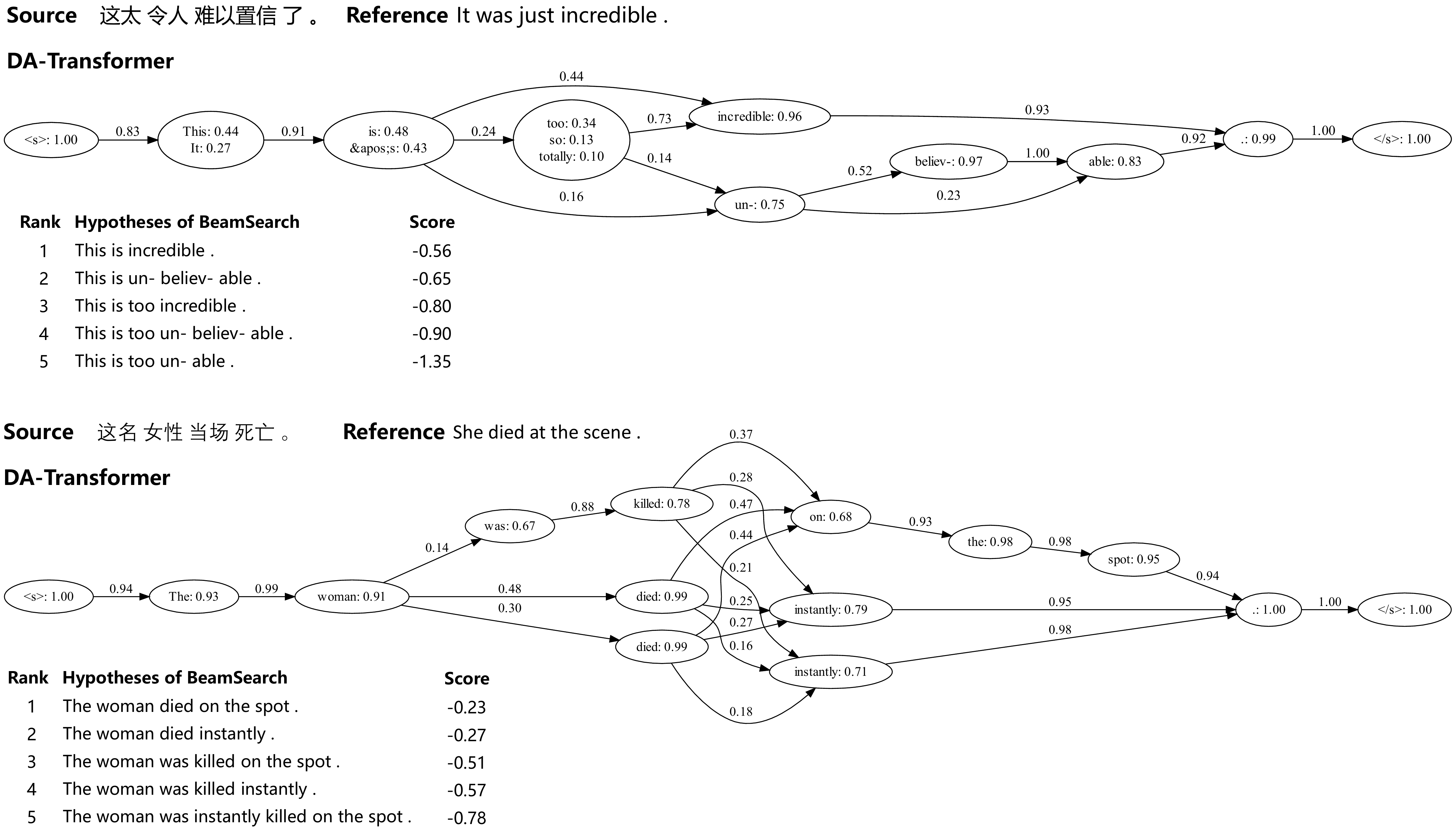}
    \vspace{-1em}
    \caption{Two test samples of WMT17 Zh-En with the \graphname and BeamSearch results.
    We show the top candidates on each vertex and the transition probabilities between vertices. We remove vertices/edges with small passing/transition probabilities for a clear presentation.
    We use the BPE tokenizer \cite{bpe2016sennrich}, where a subword prefix is marked by \textsf{-}. 
    }
    \label{fig:appendix_case}
    \vspace{0em}
\end{figure}

Two more test cases from WMT17 Zh-En are presented in Fig.\ref{fig:appendix_case}.

\label{app:more_cases}

\section{Statistics of DAGs}

\label{app:dag_stats}

For a better understanding of \methodname, we collect some statistics of predicted DAGs on WMT17 Zh-En. We use a \methodname with $\lambda=4$.

\begin{figure}[!h]
    \centering
    \includegraphics[width=0.6\linewidth]{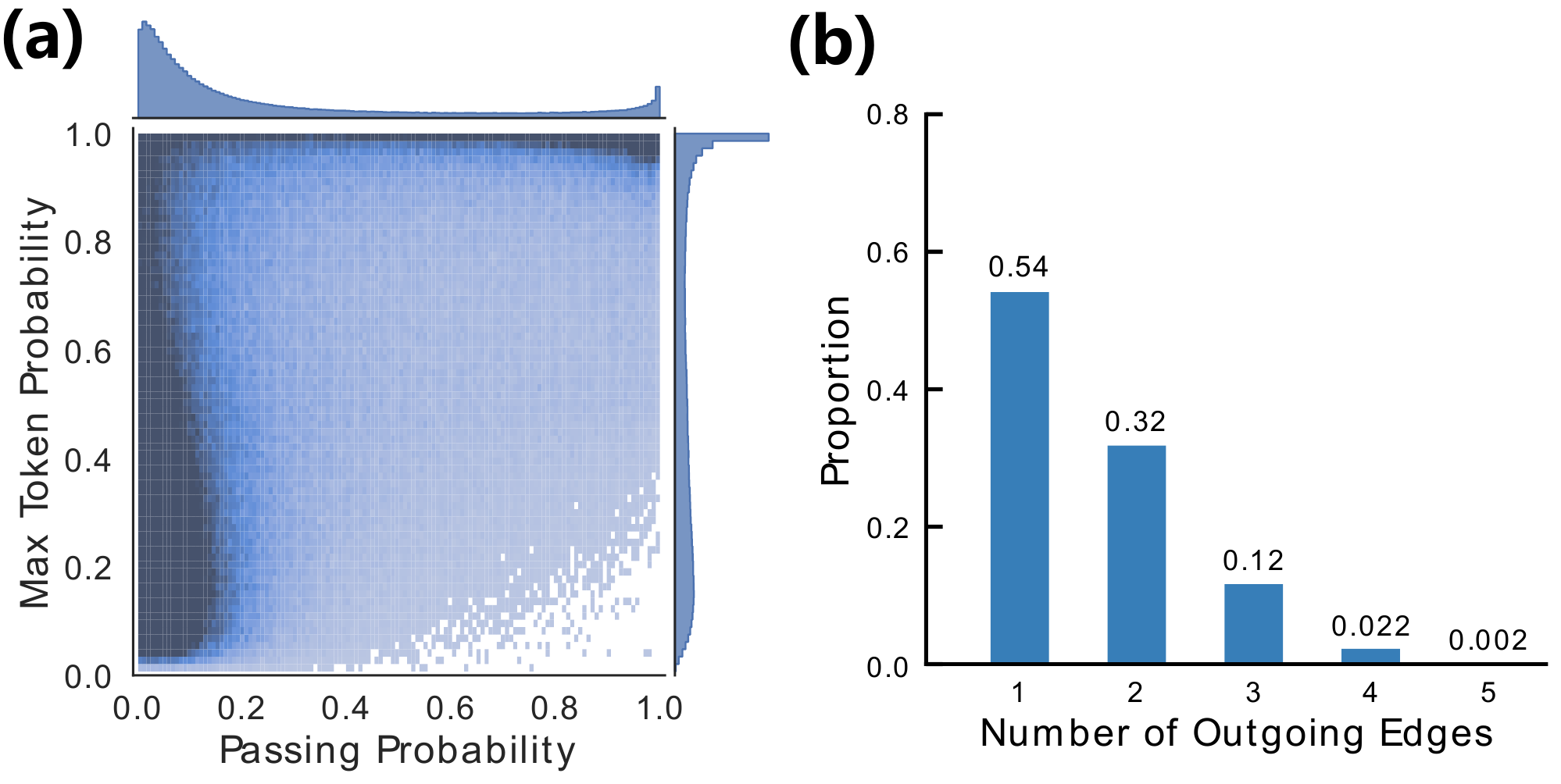}
    \vspace{-0.5em}
    \caption{Statistics of \graphnames. (a) The distribution of vertices' passing probabilities and max token probabilities. The passing probability represents how likely a vertex would appear on a randomly sampled path. The max token probability represents the probability of the most probable token on the vertex.
    (b) The distribution of numbers of vertices' outgoing edges. We only consider the most likely edges accounting for 80\% of the transition probabilities and ignore the vertices with passing probabilities smaller than 0.2. We further merge the edges that are linked to vertices predicting the same token.
    }
    \label{fig:stats}
    \vspace{-1em}
\end{figure}

In Fig.\ref{fig:stats} (a), we present the distribution of vertices with passing probability and max token probability.
We generally divide the vertices into three categories:
\vspace{-0.3em}

\begin{itemize}[leftmargin=1em]
    \setlength{\itemsep}{0ex}{
    \setlength{\parskip}{2px}{
    \item Vertices with Passing Prob $> 0.5$ (accounting for 19.6\%): They are very likely to appear in the generated translation. Since the average target length is about $\frac{1}{\lambda} = 25\%$ of the graph size, these vertices generate most of the tokens in the outputs.
    \item Vertices with Passing Prob $< 0.5$ and Max Token Prob $> 0.5$ (accounting for 40.2\%): They have high confidence in predicting tokens but do not usually appear in the translation. They may contain some rare expressions.
    \item Vertices with Passing Prob $< 0.2$ and Max Token Prob $< 0.2$ (accounting for 15.4\%): These vertices do not have specific meanings. We think the vertices are not well learned. It may be helpful if we encourage them to be more confident in generating some specific tokens.
    }
    }
\end{itemize}
\vspace{-0.3em}

In Fig.\ref{fig:stats} (b), we present the number of outgoing edges of vertices. We find that half of the vertices have only one outgoing edge, and the other half have multiple edges. The result shows that the predicted DAGs have complicated structures, which do not degenerate into chains.

%%%%%%%%%%%%%%%%%%%%%%%%%%%%%%%%%%%%%%%%%%%%%%%%%%%%%%%%%%%%%%%%%%%%%%%%%%%%%%%
%%%%%%%%%%%%%%%%%%%%%%%%%%%%%%%%%%%%%%%%%%%%%%%%%%%%%%%%%%%%%%%%%%%%%%%%%%%%%%%

\end{CJK}
\end{document}